%% file: main.tex
\documentclass{article}

\usepackage{microtype}
\usepackage{graphicx}
\usepackage{subfigure}
\usepackage{booktabs} %
\usepackage{wrapfig} %
\usepackage{enumitem}

\usepackage{hyperref}

\usepackage[accepted]{icml2024}

\usepackage{amsmath}
\usepackage{amssymb}
\usepackage{mathtools}
\usepackage{amsthm}
\usepackage[most]{tcolorbox}
\usepackage{moresize}

\usepackage[capitalize,noabbrev]{cleveref}

\theoremstyle{plain}

\theoremstyle{definition}

\theoremstyle{remark}

\usepackage[textsize=tiny]{todonotes}

\definecolor{cb_green}{RGB}{34,136,51}
\definecolor{cb_red}{RGB}{204, 51, 17}

\icmltitlerunning{A Human-Inspired Reading Agent with Gist Memory}

\begin{document}

\twocolumn[
\icmltitle{A Human-Inspired Reading Agent with Gist Memory of Very Long Contexts}

\begin{icmlauthorlist}
\icmlauthor{Kuang-Huei Lee}{gdm}
\icmlauthor{Xinyun Chen}{gdm}
\icmlauthor{Hiroki Furuta}{gdm}
\icmlauthor{John Canny}{gdm}
\icmlauthor{Ian Fischer}{gdm}
\end{icmlauthorlist}

\icmlaffiliation{gdm}{Google DeepMind}

\icmlcorrespondingauthor{Kuang-Huei Lee}{leekh@google.com}
\icmlcorrespondingauthor{Ian Fischer}{iansf@google.com}

\icmlkeywords{LLM, agent, long-context, memory, reasoning}

\vskip 0.3in
]

\printAffiliationsAndNotice{Project website and demo: \href{https://read-agent.github.io/}{\color{blue}read-agent.github.io}. Contribution statements: \Cref{appendix:contribution}.}  %

\begin{abstract}

Current Large Language Models (LLMs) are not only limited to some maximum context length, but also are not able to robustly consume long inputs.
To address these limitations, we propose ReadAgent, an LLM agent system that increases effective context length up to $20\times$ in our experiments.
Inspired by how humans interactively read long documents, we implement ReadAgent as a simple prompting system that uses the advanced language capabilities of LLMs to (1) decide what content to store together in a memory episode, (2) compress those memory episodes into short episodic memories called \emph{gist memories}, and (3) take actions to look up passages in the original text if ReadAgent needs to remind itself of relevant details to complete a task.
We evaluate ReadAgent against baselines using retrieval methods, using the original long contexts, and using the gist memories.
These evaluations are performed on three long-document reading comprehension tasks: QuALITY, NarrativeQA, and QMSum.
ReadAgent outperforms the baselines on all three tasks while extending the effective context window by $3.5-20\times$.

\end{abstract}

\input{intro}
\input{related_work}

\input{method}

\input{experiment}

\section{Conclusion}

We have presented ReadAgent, a simple interactive prompting system to mitigate the context length and context use limitations of current LLMs.
ReadAgent outperforms other strong zero-shot (i.e., not trained or finetuned on the training set) baselines across standard performance metrics.
These results demonstrate that LLMs are capable of generating compressed textual representations of long contexts that are useful for tasks that humans think are important, even without knowing those tasks ahead of time.
They also demonstrate that LLMs are capable of reasoning interactively over such compressed representations, using them to decide what information needs to be retrieved to effectively perform a known task.
ReadAgent increases the effective context length by up to $20\times$ while outperforming conventional retrieval techniques.
However, it does not give infinite context lengths, nor does it guarantee good performance when the gist memory itself is extremely long.
Future work will need to address these fundamental limitations in LLMs.

\section*{Impact Statement}
As ReadAgent is built atop LLMs, it naturally inherits their impacts and risks.
It also makes it possible to attempt to solve new problems that current LLMs cannot tackle, due to context length limitations.
It is possible that ReadAgent could cause greater harms as a consequence, just as it could improve things, depending on how it is used.
One risk that we were not able to study, but that seems particularly plausible, is of an increased tendency of the LLM to hallucinate when working with gist memories rather than full text.
Since many details are elided in the gist memories, if the model is called upon to perform some task that requires those details, it may generate them itself without giving any indication that is the case.

\section*{Acknowledgements}
The authors thank Sergey Ioffe, Rif A. Saurous, Yujin Tang, Sergio Guadarrama, Daliang Li, Felix Yu, and Rob Fergus for valuable feedback and discussion.

\bibliography{main}
\bibliographystyle{icml2024}

\newpage
\appendix
\onecolumn
\input{appendix}

\end{document}

%% file: intro.tex
\section{Introduction}
\label{sec:intro}

\begin{figure}[ht]
    \centering
    \includegraphics{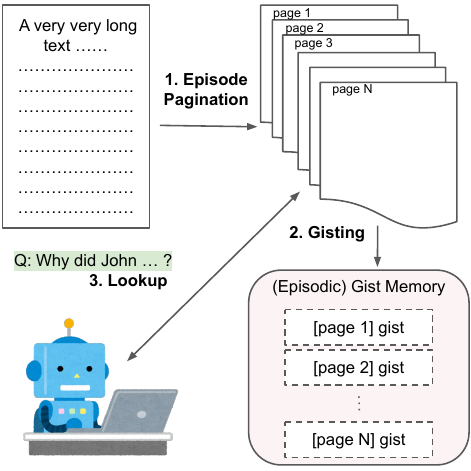}
    \caption{ReadAgent workflow.}
    \label{fig:readagent}
\end{figure}

Transformer-based Large Language Models (LLMs) are highly capable of language understanding, but the amount of text that LLMs are able to read at one time is constrained.
Not only is there an explicit context length limitation, but it has also been found that performance of LLMs tends to decline with increasingly long inputs even when they don't actually exceed the explicit context window~\cite{liu2023lost, shi2023large}.
In contrast, humans can read, understand, and reason over very long texts, such as a series of interrelated books.

We posit that an underlying reason for this gap is inherent in the differences in reading approaches.
Typically, we use LLMs to consume the exact given content word-by-word and the process is relatively passive.
On the other hand, humans read and reason over long text differently. 
First, the exact information tends to be forgotten quickly, whereas the fuzzier gist information, i.e. the substance irrespective of exact words, from past readings lasts much longer~\cite{reyna1995fuzzytrace,reyna1995fuzzyfoundation,reyna2012new}\footnote{%
    Fuzzy-trace theory~\cite{reyna1995fuzzytrace} posits that people form two types of memory representations about a past event -- verbatim and gist memories.
    Gist memories, often episodic, are fuzzy memories of past events, whereas verbatim memories contain details of past events.
    People prefer to reason with gists rather than with verbatim memories~\cite{reyna2008theory}.
}.
Second, human reading is an interactive process. 
When we need to remind ourselves of relevant details in order to complete a task, such as answering a question, we look them up in the original text.  %

We think that using the fuzzy gist memory to capture global context and attending to local details together enables humans to reason over very long context efficiently, in terms of how much information to process at once, and is also important for comprehension.
For example, if we were to infer the intention of a fictional character's specific action described on a page in a novel, besides focusing on the surrounding pages, we likely also need to understand the overall story and the character's personality from reading the whole book (see \Cref{appendix:case_study} for more analysis).

Motivated by these observations, we propose ReadAgent, an LLM agent system that handles long content inspired by the human approach.
ReadAgent is simple to implement and can be built entirely by prompting a previously-trained LLM.
As illustrated in \Cref{fig:readagent}, it takes three primary steps:
\textbf{(1) episode pagination}, where we prompt the LLM to decide where to pause in reading contiguous text; the content between pause points becomes an episode, which we refer to as \emph{pages} in this work;
\textbf{(2) memory gisting}, where we prompt the LLM to compress each page into a shorter \emph{gist} and associate the gist with a corresponding context (e.g. which page the gist was from) -- this gives the episodic \emph{gist memory};
\textbf{(3) interactive look-up}, where the LLM looks at the given task and the complete set of gists in-context, makes decision on what page(s) to look up, combines the gists with these raw pages, and solves the task.

We evaluate ReadAgent by comparing against using only the gist memory without interactive look-up, using full text for datasets that can fit in the context window, and using retrieval methods to look up pages.
ReadAgent outperforms all baselines across three challenging long-document comprehension tasks -- QuALITY, NarrativeQA and QMSum -- while increasing the effective context length significantly compared to the original LLM.
On NarrativeQA Gutenberg test set, whose average length is 71k words and whose maximum is 343k words, ReadAgent improves the LLM rating (\Cref{subsec:llm_rater}) by 12.97\% and ROUGE-L by 31.98\% over the best retrieval baseline and increases the effective context length by $\sim 20\times$.
On QuALITY, where the articles can fit in an 8K context window, ReadAgent outperforms using full text with a $3.5\times$ effective context length while saving 20.4\% on the overall number of words consumed by the LLM (\Cref{subsec:overhead}).

Finally, in \Cref{appendix:web_agent}, we adapt ReadAgent to web navigation, which is a fundamentally very-long context agent setting.
We find that ReadAgent is simple to adapt to this setting and shows promising performance.

Our primary contributions are:
\begin{itemize}[leftmargin=1em]
  \item \textbf{ReadAgent}, our human-inspired LLM agent that generates gist memories and looks up information as needed for solving tasks on long contexts (\Cref{sec:method}).
  \item Demonstration of significant performance advantages and scalability through a comprehensive experimental evaluation on challenging long-context benchmarks, comparisons against popular baselines, and analysis (\Cref{sec:exp}).
\end{itemize}

%% file: related_work.tex
\section{Related Work}
\label{sec:related_work}

\paragraph{Long-Context LLMs}
The most direct way to improve LLM long-context performance is to train or fine-tune LLMs with longer context windows~\citep{beltagy2020longformer,zaheer2020bigbird,guo2022longt5,ainslie2023colt5,10.1145/3530811,chen2023longlora}.
Another approach is to explore new architectures or efficient implementations of the Transformer~\citep{vaswani2017attention} attention layers to reduce the need of long-context fine-tuning~\citep{chen2023extending,press2022train,xiao2023efficient,jin2024llm,han2023lm}.
However, LLM performance tends to decline with increasingly long inputs even when they don't exceed the specified context length~\cite{liu2023lost}.
LLM performance is also shown to be sensitive to distracting information in the context~\cite{shi2023large}.
Thus, the effective context length could be shorter than the explicit limit.
Our approach is complimentary to these approaches, scaling the effective context length of the underlying model while reducing the amount of distracting information in context, and requiring neither architectural changes nor training.

\paragraph{Retrieval}
Retrieval Augmented Generation (RAG) techniques~\citep{chen2017reading,dinan2019wizard,lewis2020retrieval,izacard2021leveraging,wu2022memorizing,park2023generative,zhong2023memorybank} allow an LLM to query task-relevant information from a large database of documents or document pieces.
Our work implements a form of retrieval by reasoning over a contextualized gist memory, all with zero-shot LLM prompting.
This rethinking of retrieval directly leverages the strength and flexibility of LLM language understanding to reason about which documents to retrieve.
Our approach is well-suited to densely-correlated long-document pieces, such as a series of books or a conversation history, but the database cannot scale arbitrarily, since the size of the gist memory is limited by the LLM's context length, and the gist memory's length correlates with the size of the database.
In contrast, conventional retrieval approaches can handle larger databsases than our approach.
In this work, we compare against retrieval systems that use exactly the same set of documents as our approach.

\paragraph{LLM Agents for Long Texts}
LLMs can be used as agents to interactively handle very long texts.
WebGPT~\cite{nakano2021webgpt} and WebShop~\cite{yao2022webshop} learn browsing actions to search for the requested answer on the internet, despite not being designed to understand long documents.
The PEARL~\cite{sun2023pearl} system proposes action plans for better long-document comprehension through iterative prompting; \citet{yuan2020interactive} explicitly learns RL agents for similar purposes.
Self-note~\cite{lanchantin2023learning} amortizes reasoning steps and interleaves intermediate notes with the original documents to improve reasoning.
\citet{yang2022re3} generates long outputs through iterative reasoning.
However, these methods cannot address long input texts that exceed the LLM's context length.
Similar to this work, MemWalker~\cite{chen2023walking} also reads long documents interactively through iterative prompting.
It traverses a tree of different levels of summaries to search for task-related information.
However, the hierarchical summary structure makes it difficult to reason over related but distant information at the same granularity (see \Cref{appendix:memwalker} for more discussion).

%% file: method.tex
\vspace{-1mm}
\section{ReadAgent}
\label{sec:method}

\Cref{fig:readagent} shows an overview of ReadAgent, which we describe in detail below.
Note that the prompts presented in this section are examples, which may need to change according to the target task. 
We release the prompts for each task on \href{https://read-agent.github.io/}{\color{blue}read-agent.github.io}.
Please also refer to \Cref{appendix:prompt_design} for the prompt design details.

\subsection{Gist Memory}
\label{subsec:gist_mem}

A \emph{gist memory} is an ordered collection of short gists of chunks of text from the original long context.
Building a gist memory has two steps: \emph{pagination} and \emph{memory gisting}, described in turn below.

\paragraph{Episode Pagination}
When ReadAgent reads through a long text, it makes decisions on what content to store together in a memory episode by choosing where to pause reading.
At each step, we provide the LLM some text that begins from the previous pause point and ends when it reaches a \texttt{max\_words} limit.
We prompt the LLM to choose which point between paragraphs would be a natural point to pause, and then treat the content between the previous and current pause points as an episode, which we also refer as a \emph{page}.
This is \emph{episode pagination}, which we implement with the following prompt.

As shown in the prompt, possible pause points are inserted between paragraphs as numbered tags (e.g. $\langle13\rangle$), making this a multiple choice question for the LLM.
We only start inserting these numbered tags after a \texttt{min\_words} threshold to make sure that each page has at least \texttt{min\_words}.

\vspace{1mm}

\begin{tcolorbox}[breakable,title=Example Pagination Prompt]
\footnotesize
You are given a passage that is taken from a larger text (article, book, ...) and some numbered labels between the paragraphs in the passage.
\vspace{1mm}

Numbered labels are in angle brackets. For example, if the label number is 19, it shows as $\langle19\rangle$ in text.
\vspace{1mm}

Please choose a label where it is natural to break reading.
\vspace{1mm}

The label can be a scene transition, the end of a dialogue, the end of an argument, a narrative transition, etc.
\vspace{1mm}

Please answer with the break point label and explain.
\vspace{1mm}

For example, if $\langle57\rangle$ is a good point to break, answer with ``Break point: $\langle57\rangle$\textbackslash\!n Because ...''
\vspace{1mm}

Passage:

\{...\}\newline
\{PARAGRAPH 5 TEXT\}\newline
$\langle5\rangle$\newline
\{PARAGRAPH 6 TEXT\}\newline
$\langle6\rangle$\newline
\{PARAGRAPH 7 TEXT\}\newline
\{...\}
\end{tcolorbox}

\vspace{-3mm}
\paragraph{Memory Gisting}
For each \emph{page}, we prompt the LLM to shorten the exact content into a \emph{gist}, or summary, as follows.

\begin{tcolorbox}[title=Example Gisting Prompt]
\footnotesize
Please shorten the following passage.
\vspace{1mm}

Just give me a shortened version. DO NOT explain your reason.
\vspace{1mm}

Passage: 

\{PAGE TEXT\}
\end{tcolorbox}

We subsequently prepend a page tag to each gist (e.g. ``$\langle\textrm{Page 2}\rangle$\!\textbackslash\!n\{GIST CONTENT\}'') to contextualize it (indicate where the gist was from), and then concatenate all gists.
This gives us the gist memory.
We use the word ``shorten'' in the prompt to generate these summarizing gists as it tends to help preserve the narrative flow, making it more natural to concatenate.
Using the word ``summarize'' tended to produce a restructured summary in our experiments.

The original page size is a key factor for how compressed the gist is.
Let's say the smallest unit of text that we consider is a paragraph.
Intuitively, a paragraph likely has some amount of mutual information with its neighbors.
Thus, the larger chunk of text we group together, the more duplicated information we can remove.
Empirically, compressing larger chunks of text with LLMs also tends to remove more details, which could affect performance.
We control the page size by changing \texttt{min\_words} and \texttt{max\_words} in pagination.
This trade-off is studied in \Cref{subsec:ablation}.

\vspace{-1mm}
\subsection{Interactive Look-Up and Response}
\label{subsec:lookup}
For a given task about a long document, we want ReadAgent to take actions to look up relevant details in the original text in addition to using its gist memory.
As the gist memories are contextualized with page numbers, we simply prompt the LLM to answer which page(s) it would like to look up and read again given the specific task.
In the following we discuss two look-up strategies: looking up all pages at once in parallel (\textbf{ReadAgent-P}) and sequentially looking up one page at a time (\textbf{ReadAgent-S}).

\paragraph{ReadAgent-P} As in the following example prompt for question-answering, typically we give it a maximum number of pages that it can look up but also instruct it to use as few pages as possible to avoid unnecessary computational overhead and distracting information.
The following prompt shows parallel look-up, where the model requests multiple pages in response to a single prompt.

\begin{tcolorbox}[breakable,title=Example Parallel Lookup Prompt (ReadAgent-P)]
\footnotesize
The following text is what you remember from reading an article and a multiple choice question related to it.
\vspace{1mm}

You may read 1 to 5 page(s) of the article again to refresh your memory to prepare yourself for the question.
\vspace{1mm}

Please respond with which page(s) you would like to read.
\vspace{1mm}

For example, if you only need to read Page 8, respond with ``I want to look up Page [8] to ...''; if you would like to read Page 7 and 12, respond with ``I want to look up Page [7, 12] to ...''; if you would like to read Page 2, 3, 7, 15 and 18, respond with ``I want to look up Page [2, 3, 7, 15, 18] to ...''.
\vspace{1mm}

DO NOT select more pages if you don't need to.
\vspace{1mm}

You don't need to answer the question yet.
\vspace{1mm}

Text:

\{GIST MEMORY\}
\vspace{1mm}

Question:

\{QUESTION\}

\end{tcolorbox}

The selected raw pages replace the gist(s) at the corresponding positions in memory, preserving the overall narrative flow.
Then we prompt the LLM again with the task and the updated memory and ask it to solve the task (see example prompts in \Cref{appendix:prompt_design}).

\paragraph{ReadAgent-S} We also study the sequential look-up strategy, where the model requests one page at a time, up to some maximum number of pages.
In sequential look-up, the model gets to see the previously expanded pages before deciding which page to expand.
This gives the model access to more information than parallel look-up, so we might expect it to perform better in some situations.
However, the larger number of interactions with the model increases the computational cost, so sequential look-up should only be used on tasks where it provides clear benefits.

\vspace{1mm}
\begin{tcolorbox}[breakable,title=Example Sequential Lookup Prompt (ReadAgent-S)]
\footnotesize
The following text is what you remember from reading a meeting transcript, followed by a question about the transcript.

\vspace{1mm}
You may read multiple pages of the transcript again to refresh your memory and prepare to answer the question.

\vspace{1mm}
Each page that you re-read can significantly improve your chance of answering the question correctly.

\vspace{1mm}
Please specify a SINGLE page you would like to read again or say "STOP".

\vspace{1mm}
To read a page again, respond with ``Page \$PAGE\_NUM'', replacing \$PAGE\_NUM with the target page number.

\vspace{1mm}
You can only specify a SINGLE page in your response at this time.

\vspace{1mm}
To stop, simply say ``STOP''.
DO NOT answer the question in your response.

\vspace{1mm}
Text:

\{GISTS WITH IN-LINE EXPANDED PAGES\}

\vspace{1mm}
Pages re-read already (DO NOT ask to read them again):

\{LIST OF PAGE NUMBERS ALREADY READ\}

\vspace{1mm}
Question:

\{QUESTION\}

\vspace{1mm}
Specify a SINGLE page to read again, or say STOP:
\end{tcolorbox}

\subsection{Computational Trade-offs and Scalability}
\label{subsec:overhead}

Episode pagination, memory gisting and interactive look-ups require iterative inference.
As we show in the following, the additional cost is bounded linearly by a small factor, making our approach scale well with input length.

\textbf{Pagination}
In theory, an LLM could read a document and directly provide the pagination in a single pass, so the minimum number of words the LLM must process is the length of the document.
Our pagination algorithm splits the document into chunks of at most \texttt{max\textunderscore words}, and then guarantees that at least \texttt{min\textunderscore words} are consumed at each step.
Thus, the ratio $\frac{\text{\texttt{max\textunderscore words}}}{\text{\texttt{min\textunderscore words}}}$ gives an upper bound on how many times the word length of the document the LLM must process using our algorithm.
\textbf{Gisting:}
Memory gisting is one additional pass of the raw input words, since each page is gisted independently.
\textbf{Look-ups:}
Parallel look-ups are conditioned on gists instead of the full text, and thus will be much shorter than one pass of the raw input words.
Each step of a sequential look-up is similar to parallel look-ups and the overall cost is capped with the maximum number of look-ups allowed.
\textbf{Response:}
Finally, answering is also similar to parallel look-ups.
There is additional overhead from the prompt templates, of course.

On the other hand, as generating gists is an one-time effort while the look-up and response steps operate mostly on gists that are much shorter than the original text, the one-time effort can be amortized when the same context is reused for multiple tasks.
Thus, in such settings, ReadAgent can reduce the overall number of tokens to process. 
In particular, directly answering from the original QuALITY dev set (230 articles and 2086 questions) is 8,708,434 words consumed by the LLM, whereas using ReadAgent with 1-page lookup is 6,499,856 words (25.4\% saving), up-to-2-page lookup is 6,933,357 words (20.4\% saving), and up-to-5-page lookup is 7,503,084 words (13.8\% saving). 
We can expect the savings to be more significant with higher compression rate and more downstream tasks.

\subsection{ReadAgent Variants}
\label{subsec:variants}
In \Cref{appendix:variants}, we discuss variants of ReadAgent that can be useful in different problem settings, including when the target task is known prior to reading the long document.
In \Cref{appendix:web_agent}, we describe adapting ReadAgent to work in the web navigation setting.

%% file: experiment.tex
\section{Experiments}
\label{sec:exp}
We evaluate ReadAgent's long-document reading comprehension ability on three long-context question-answering challenges: 
QuALITY \cite{pang2022quality}, NarrativeQA \cite{kovcisky2018narrativeqa} and QMSum \cite{zhong2021qmsum}. 
Although ReadAgent does not require any model training, we develop the proposed method on the training sets and test on the validation, test and/or development sets to avoid any risk of overfitting system hyperparameters. 

In this work, we primarily use the instruction-tuned PaLM 2-L~\cite{anil2023palm} for our experiments and evaluation.
The context length of PaLM 2-L is 8K tokens.
Details of the model can be found in \citet{anil2023palm}.
Additionally, we provide GPT-3.5\footnote{http://openai.com/api/} results in \Cref{appendix:gpt}, and experimental results on the web navigation setting in \Cref{appendix:web_agent}.

One important performance measure of the techniques considered here is the \textbf{compression rate} (\textbf{CR}).
As we want to measure the longest LLM context length that ReadAgent requires versus full-context length, we define this as $\text{CR} \equiv 100 * (1 - \frac{\text{word-count}(\text{in-context text})}{\text{word-count}(\text{full-context text})})$ at the final response query, where the in-context text (gists and retrieved pages) length is the longest among all inference steps.

\subsection{LLM Raters}
\label{subsec:llm_rater}

NarrativeQA and QMSum both have one or more free-form reference responses.
They are typically evaluated using syntactic matching metrics such as ROUGE~\citep{lin2004rouge} F-Measure.
We additionally evaluate these datasets using an automatic LLM Rater as an alternative to human evaluation similar to~\citet{peng2023instruction,vicuna2023,zheng2023judging,chiang-lee-2023-large}.

In our implementation, we prompt the LLM to look at the question or instruction and compare the model's answer to the reference answer.
The ``Strict LLM Rater Prompt'' shown below is for judging whether there is an exact match, and the ``Permissive LLM Rater Prompt'' is for judging whether there is an exact match or a partial match. 
We apply both prompts to all model responses.
If either rater decides there is an exact match, we count it as an exact match.
If the strict rater is negative but the permissive rater detects a partial match, we count it as a partial match.
Otherwise, it's not a match.
In the case that there are multiple reference answers, the response is compared against each reference answer in turn, and the highest rating is returned.

\begin{tcolorbox}[title=Strict LLM Rater Prompt]
\footnotesize
After reading some text, John was given the following question about the text:

\vspace{1mm}
\{QUESTION TEXT\}

\vspace{1mm}
John's answer to the question was:

\vspace{1mm}
\{MODEL RESPONSE TEXT\}

\vspace{1mm}
The ground truth answer was:

\vspace{1mm}
\{REFERENCE RESPONSE TEXT\}

\vspace{1mm}
Does John's answer agree with the ground truth answer? Please answer YES or NO.
\end{tcolorbox}

\begin{tcolorbox}[title=Permissive LLM Rater Prompt]
\footnotesize
After reading some text, John was given the following question about the text:

\vspace{1mm}
\{QUESTION TEXT\}

\vspace{1mm}
John's answer to the question was:

\vspace{1mm}
\{MODEL RESPONSE TEXT\}

\vspace{1mm}
The ground truth answer was:

\vspace{1mm}
\{REFERENCE RESPONSE TEXT\}

\vspace{1mm}
Does John's answer agree with the ground truth answer? Please answer ``Yes'', ``Yes, partially'', or ``No''.
If John's response has any overlap with the ground truth answer, answer ``Yes, partially''.
If John's response contains the ground truth answer, answer ``Yes''.
If John's response is more specific than the ground truth answer, answer ``Yes''.
\end{tcolorbox}

Based on these raters, we define two different scores: \textbf{LLM-Rating-1} (\textbf{LR-1}) is a strict evaluation score, where we count the percentage of exact matches over all examples; \textbf{LLM-Rating-2} (\textbf{LR-2}) is permissive, where we count the percentage of exact and partial matches.

\subsection{Baseline Methods}
\label{subsec:baseline}

\paragraph{Retrieval-Augmented Generation (RAG)}
As discussed in~\Cref{sec:related_work}, RAG~\cite{lewis2020retrieval} is a popular approach to extend access to a large amount of text beyond what can fit in the LLM context window.
In this paper we compare ReadAgent to RAG baselines using conventional retrieval methods to find relevant ``pages'' in a long text, where we reuse the pages generated by ReadAgent.
We consider two relevance methods: Okapi BM25~\cite{robertson2009probabilistic} and neural retrieval based on the Gemini API embedding model (models/embedding-001)\footnote{https://ai.google.dev/models/gemini}.
The neural retrieval relevance score is defined as the dot product between the question embedding vector and each page (or gist memory embedding vector in the case of NarrativeQA, see \Cref{sec:narrativeqa}).
For reading comprehension tasks, the pages are ranked by relevance to each question, and we prompt the LLM to look at the top-$k$ pages as context for answering the question.
In most retrieval settings, the database of documents is quite large, which makes the retrieval task more challenging.
In our setting, ReadAgent and retrieval methods all use a per-document database, rather than per-dataset.
For example, in QuALITY, there are hundreds of articles, each with multiple questions.
The database for retrieval in each question is only the extracted pages from the corresponding article (typically less than 20 pages), rather than the thousands of pages from the entire dataset.

\paragraph{Full or Truncated Text Content}
The maximum length of QuALITY dev articles is $\sim$6,000 words, which can fit into the PaLM 2-L context window.
This allows us to evaluate ReadAgent against directly using the full long document for long-context reading comprehension.
The maximum length of QMSum is over 26,000 words.
Consequently, we choose to truncate the text to close to the context window limit (6,000 words for PaLM 2-L experiments) to ensure that the truncated text fits in the LLM's context, though this would generally be a weaker baseline.
Finally, since the average length of NarrativeQA documents significantly exceeds the context window, it is less meaningful to perform the truncated-context comparison.

\paragraph{Gist Memory}
We can also attempt to solve the given task by reasoning directly over the gist memory.
Doing so helps us understand not only the importance of interactive look-up but also how using the LLM-compressed information alone compares to the full content and retrieval baselines.

\subsection{Long-Context Reading Comprehension}
\label{subsec:long_reading_results}

\subsubsection{QuALITY}
\label{sec:quality}

QuALITY~\cite{pang2022quality} is a four-way multiple choice question answering challenge with text data from several different sources.
QuALITY is evaluated using accuracy, with 25\% corresponding to chance performance.

\begin{figure}[tbh]
    \centering
    \includegraphics[width=0.45\textwidth]{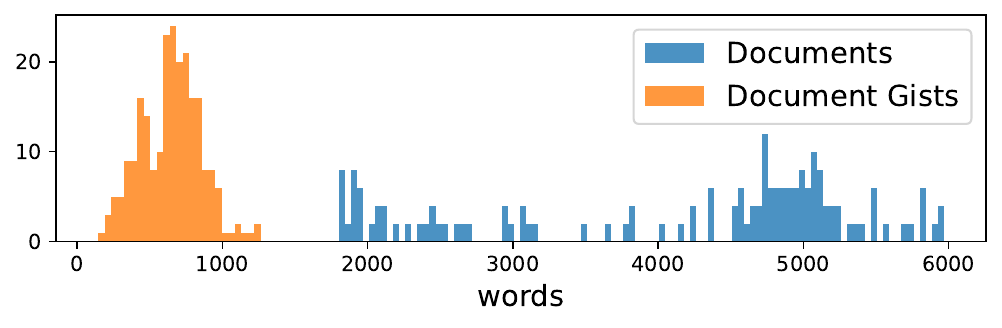}
    \vspace{-0.3cm}
    \caption{Histogram of QuALITY document and gist word counts.}
    \label{fig:quality_stats}
\end{figure}

The dev set has an average length of 4,122 words and a maximum of 5,967. 
The gist memory has an average length of 650 words and a maximum of 1,264. 
\Cref{fig:quality_stats} shows the word statistics for the original text and the gists.
The compression rate of the gists is 85.53\%.
See \Cref{appendix:details} for QuALITY pagination hyperparameters.

\Cref{tab:quality_results} shows the experimental results on QuALITY.
The performance of ReadAgent increases as we increase the maximum number of pages allowed for look-up.
ReadAgent-P (Look up 1-6 pages) achieves 86.91\% and ReadAgent-S (Look up 1-6 pages) achieves 87.17\% in accuracy.
Notably, starting from ReadAgent (Look up 1-2 pages), it outperforms all baselines methods including using the full original text, which could have been an upper bound on the performance -- every other method reduces the amount of text the LLM considers before generating its response.
However, this is not a surprising result.
Prior work shows that current LLMs are not able to effectively use the full long context window~\cite{liu2023lost}, potentially due to training data sparsity, and distracting information can also reduce performance~\cite{shi2023large, weston2023system}.
The corresponding compression rate of ReadAgent (Look up 1-2 pages) is 72.17\%, meaning that $\sim 3.5\times$ as many tokens can fit in the context window after gisting.

\begin{table}[htb]
    \small
    \centering
    \begin{tabular}{l|c|c}
        \hline
        Method & CR (\# LU) & Accuracy \\
        \hline
        \multicolumn{3}{l}{BM25 Retrieval} \\
        \hline
        \hspace{3mm} Top-1 & 89.27\% (1) & 70.34\% $\pm$ 0.06 \\
        \hspace{3mm} Top-2 & 78.96\% (2) & 79.05\% $\pm$ 0.05 \\
        \hspace{3mm} Top-3 & 68.50\% (3) & 82.65\% $\pm$ 0.05 \\
        \hspace{3mm} Top-4 & 58.57\% (4) & 84.42\% $\pm$ 0.13 \\
        \hline
        \multicolumn{3}{l}{Neural Retrieval with Gemini API} \\
        \hline
        \hspace{3mm} Top-1 & 89.91\% (1) & 71.32\% $\pm$ 0.19 \\
        \hspace{3mm} Top-2 & 80.08\% (2) & 79.02\% $\pm$ 0.10 \\
        \hspace{3mm} Top-3 & 70.28\% (3) & 83.41\% $\pm$ 0.10 \\
        \hspace{3mm} Top-4 & 60.68\% (4) & 84.88\% $\pm$ 0.03 \\
        \hline
        Full Raw Content & 0\% & 85.83\% $\pm$ 0.19 \\
        \hline
        \textbf{GistMem} & 85.53\% & 77.52\% $\pm$ 0.13 \\
        \hline
        \multicolumn{3}{l}{\textbf{ReadAgent-P}} \\
        \hline
        \hspace{3mm} Look up 1 pg    & 76.00\% (1.0) & 84.13\% $\pm$ 0.10 \\
        \hspace{3mm} Look up 1-2 pgs & 72.17\% (1.6) & 86.16\% $\pm$ 0.12 \\
        \hspace{3mm} Look up 1-3 pgs & 69.36\% (2.0) & 86.59\% $\pm$ 0.10 \\
        \hspace{3mm} Look up 1-4 pgs & 67.73\% (2.2) & 86.86\% $\pm$ 0.00\\
        \hspace{3mm} Look up 1-5 pgs & 66.45\% (2.3) & 86.83\% $\pm$ 0.10 \\
        \hspace{3mm} Look up 1-6 pgs & 64.75\% (2.5) & \textbf{86.91\%} $\pm$ 0.08 \\
        \hline
        \textbf{ReadAgent-S} 1-6 pgs & 58.53\% (3.2) & \textbf{87.17\%} $\pm$ 0.18 \\
        \hline
    \end{tabular}
    \caption{
    QuALITY results on the dev set of 230 docs and 2086 questions using PaLM 2-L. 
    \textbf{CR} is the compression rate.
    \textbf{\# LU} is the number of lookups.
    We report means and standard deviations across 3 runs, except where inconsequential (CR and \# LU).
    } 
    \label{tab:quality_results}
    \vspace{-1em}
\end{table}

\begin{table*}[t!]
    \scriptsize
    \centering
    \begin{tabular}{l|c|c|c|c|c|c||c|c|c|c|c|c}
        \hline
         & \multicolumn{6}{c||}{Gutenberg Validation (58 docs \& 1743 questions)} &  \multicolumn{6}{c}{Gutenberg Test (177 docs \& 5207 questions)} \\
        \hline
        Method & CR (\# LU) & LR-1 & LR-2 & R-1 & R-2 & R-L & CR (\# LU) & LR-1 & LR-2 & R-1 & R-2 & R-L \\
        \hline
        \multicolumn{13}{l}{BM25 Retrieval} \\
        \hline
        \hspace{2mm} Top-1 & 97.63\% (1) & 39.01\% & 50.14\% & 0.166 & 0.061 & 0.156 & 97.42\% (1) & 43.5\% & 55.33\% & 0.176 & 0.065 & 0.165 \\
        \hspace{2mm} Top-2 & 95.24\% (2) & 49.34\% & 60.76\% & 0.203 & 0.079 & 0.191 & 94.80\% (2) &  51.70\% & 64.53\% & 0.206 & 0.082 & 0.194 \\
        \hspace{2mm} Top-3 & 93.34\% (3) & 52.73\% & 63.68\% & 0.208 & 0.080 & 0.195 & 93.02\% (3) & 52.97\% & 66.03\% & 0.210 & 0.083 & 0.197 \\
        \hspace{2mm} Top-4 & 92.47\% (4) & 53.59\% & 64.26\% & 0.211 & 0.082 & 0.197 & 92.27\% (4) & 53.60\% & 66.16\% & 0.210 & 0.084 & 0.197 \\
        \hline
        \multicolumn{13}{l}{Neural Retrieval with Gemini API} \\
        \hline
        \hspace{2mm} Top-1 & 98.19\% (1) & 34.25\% & 46.53\% & 0.146 & 0.051 & 0.134 & 98.14\% (1) & 36.47\% & 47.8\% & 0.150 & 0.054 & 0.140 \\
        \hspace{2mm} Top-2 & 96.30\% (2) & 44.69\% & 54.96\% & 0.180 & 0.069 & 0.167 & 96.15\% (2) & 44.48\% & 56.17\% & 0.182 & 0.070 & 0.170 \\
        \hspace{2mm} Top-3 & 94.62\% (3) & 46.24\% & 57.31\% & 0.191 & 0.077 & 0.178 & 94.42\% (3) & 48.97\% & 60.73\% & 0.195 & 0.076 & 0.183 \\
        \hspace{2mm} Top-4 & 93.45\% (4) & 48.59\% & 59.21\% & 0.196 & 0.079 & 0.184 & 93.25\% (4) & 50.62\% & 62.05\% & 0.203 & 0.080 & 0.191 \\
        \hline
        \textbf{GistMem} & 96.89\% & 55.31\% & 68.22\% & 0.233 & 0.091 & 0.218 & 96.80\% & 55.79\% & 71.19\% & 0.231 & 0.092 & 0.217 \\
        \hline
        \multicolumn{13}{l}{{\textbf{ReadAgent-P}}} \\
        \hline
        \hspace{2mm} Look up 1 pg & 95.15\% (0.94) & 58.92\% & 71.89\% & \bf{0.244} & \bf{0.101} & \bf{0.230} & 94.84\% (0.93) & 59.98\% & \bf{73.23}\% & \bf{0.240} & \bf{0.098} & \bf{0.226} \\
        \hspace{2mm} Look up 1-2 pgs & 94.79\% (1.23) & \bf{59.84\%} & \bf{72.29\%} & 0.239 & 0.098 & 0.224 & 94.36\% (1.34) & 59.19\% & 72.65\% & 0.231 & 0.091 & 0.218 \\
        \hspace{2mm} Look up 1-3 pgs & 94.39\% (1.50) & \bf{59.84\%} & 71.89\% & 0.240 & 0.098 & 0.226 & 94.03\% (1.61) & 59.63\% & 72.84\% & 0.230 & 0.093 & 0.217 \\
        \hline
        \textbf{ReadAgent-S} 1-2 pgs & 94.35\% (1.38) & 57.89\% & 71.14\% & 0.239 & 0.097 & 0.225 & 93.86\% (1.46) & 60.48\% & 72.48\% & 0.232 & 0.095 & 0.219 \\
        \textbf{ReadAgent-S} 1-3 pgs & 94.08\% (1.57) & 58.52\% & 71.49\% & 0.242 & 0.098 & 0.229 & 93.67\% (1.57) & \bf{60.55\%} & 72.79\% & 0.231 & 0.095 & 0.219\\
        \hline
        \hline
        & \multicolumn{6}{c||}{Movie Validation (57 docs \& 1699 questions)} & \multicolumn{6}{c}{Movie Test (172 docs \& 5139 questions)} \\
        \hline
        \multicolumn{13}{l}{BM25 Retrieval} \\
        \hline
        \hspace{2mm} Top-1 & 97.07\% (1) & 32.67\% & 42.61\% & 0.156 & 0.058 & 0.144 & 96.61\% (1) & 33.64\% & 43.34\% & 0.154 & 0.054 & 0.143 \\
        \hspace{2mm} Top-2 & 94.12\% (2) & 39.97\% & 50.21\% & 0.187 & 0.070 & 0.174 & 93.81\% (2) & 42.50\% & 53.05\% & 0.191 & 0.072 & 0.178 \\
        \hspace{2mm} Top-3 & 91.18\% (3) & 43.61\% & 53.91\% & 0.198 & 0.077 & 0.185 & 91.00\% (3) & 46.97\% & 57.52\% & 0.207 & 0.080 & 0.193 \\
        \hspace{2mm} Top-4 & 88.24\% (4) & 46.85\% & 57.62\% & 0.210 & 0.084 & 0.198 & 88.19\% (4) & 50.18\% & 60.13\% & 0.217 & 0.085 & 0.202 \\
        \hline
        \multicolumn{13}{l}{Neural Retrieval with Gemini API} \\
        \hline
        \hspace{2mm} Top-1 & 97.07\% (1) & 32.02\% & 41.44\% & 0.153 & 0.053 & 0.142 & 96.67\% (1) & 37.24\% & 46.22\% & 0.130 & 0.043 & 0.118 \\
        \hspace{2mm} Top-2 & 94.19\% (2) & 43.20\% & 51.38\% & 0.160 & 0.057 & 0.148 & 93.90\% (2) & 46.49\% & 54.60\% & 0.164 & 0.061 & 0.151 \\
        \hspace{2mm} Top-3 & 91.29\% (3) & 47.56\% & 56.21\% & 0.176 & 0.064 & 0.163 & 91.14\% (3) & 50.69\% & 58.92\% & 0.186 & 0.071 & 0.172 \\
        \hspace{2mm} Top-4 & 88.38\% (4) & 49.09\% & 59.33\% & 0.193 & 0.075 & 0.180 & 88.36\% (4) & 52.13\% & 59.41\% & 0.184 & 0.072 & 0.171 \\
        \hline
        \textbf{GistMem} & 92.09\% & 52.56\% & 64.39\% & 0.242 & 0.103 & 0.227 & 91.98\% & 54.68\% & 64.00\% & 0.248 & 0.105 & 0.234 \\
        \hline
        \multicolumn{13}{l}{{\textbf{ReadAgent-P}}} \\
        \hline
        \hspace{2mm} Look up 1 pg & 89.20\% (0.99) & 53.38\% & 65.57\% & \bf{0.247} & \bf{0.106} & \bf{0.233} & 89.22\% (0.98) & 57.68\% & 68.01\% & \bf{0.274} & \bf{0.116} & \bf{0.260} \\
        \hspace{2mm} Look up 1-2 pgs & 87.68\% (1.52) & 54.62\% & 65.63\% & 0.238 & 0.098 & 0.223 & 88.10\% (1.39) & 58.24\% & 68.81\% & 0.270 & 0.115 & 0.255 \\
        \hspace{2mm} Look up 1-3 pgs & 86.57\% (1.91) & 54.91\% & 65.86\% & 0.241 & 0.099 & 0.225 & 86.73\% (1.89) & 58.82\% & 69.12\% & 0.272 & \bf{0.116} & 0.257 \\
        \hline
        \textbf{ReadAgent-S} 1-2 pgs & 86.36\% (1.98) & 59.33\% & 68.28\% & 0.203 & 0.082 & 0.188 & 85.92\% (1.98)  & 63.33\%  & 72.06\%  & 0.214  & 0.086  & 0.199\\
        \textbf{ReadAgent-S} 1-3 pgs & 83.56\% (2.95) & \bf{59.45\%} & \bf{68.81\%} & 0.210 & 0.087 & 0.195 & 83.18\% (2.95) & \bf{64.53\%} & \bf{73.06\%} & 0.217 & 0.090  & 0.202 \\
        \hline
    \end{tabular}
    \caption{
    NarrativeQA results (PaLM 2-L). 
    \textbf{R-1}, \textbf{R-2}, and \textbf{R-L} are ROUGE F-Measures.
    \textbf{LR-1}, and \textbf{LR-2} are LLM-Ratings.
    }
    \label{tab:narrativeqa_results}
    \vspace{-1em}
\end{table*}

\begin{figure}[tb]
    \centering
    \includegraphics[width=0.45\textwidth]{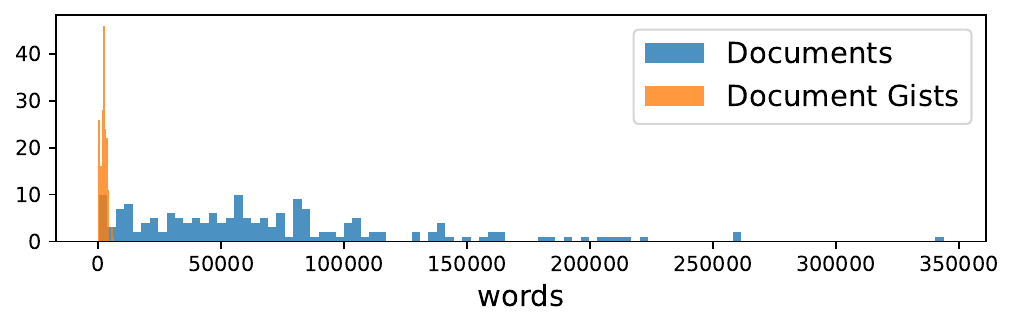}
    \caption{Histogram of NarrativeQA (Gutenberg) test set word counts for the original text and the gists.}
    \label{fig:narrativeqa_gutenberg_stats}
    \vspace{-2mm}
\end{figure}

\begin{figure}[tb]
    \centering
    \includegraphics[width=0.45\textwidth]{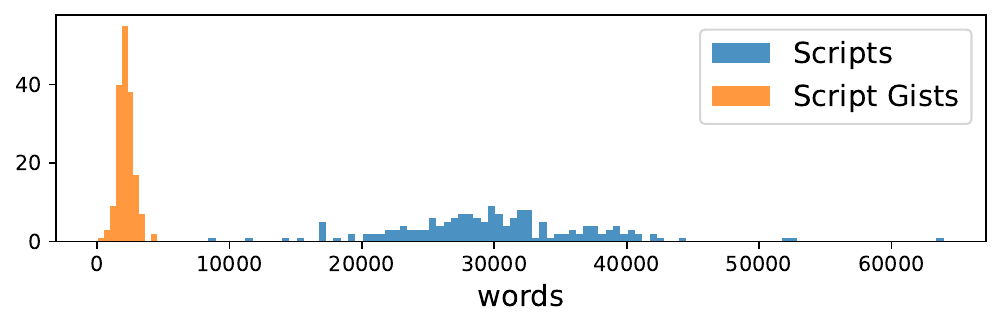}
    \caption{Histogram of NarrativeQA (movie) test set word counts for the original text and the gists.}
    \label{fig:narrativeqa_movie_stats}
    \vspace{-2mm}
\end{figure}

\subsubsection{NarrativeQA}
\label{sec:narrativeqa}
NarrativeQA~\cite{kovcisky2018narrativeqa} has the longest context length on average among the three reading comprehension datasets we choose.
The dataset is divided into books (Gutenberg) and move scripts.
The Gutenberg test set have 70,619 words on average, and the maximum is 343,910 words; the movie scripts test set have 29,963 on average, and the maximum is 63,957 words.
As the reference answers are free-form, we evaluate based on ROUGE~\cite{lin2004rouge} and the LLM Ratings (\Cref{subsec:llm_rater}).
The original main texts are replaced with the HTML-stripped version from SCROLLS~\cite{shaham2022scrolls}.

Because of the length of NarrativeQA articles, in order to fit the gists into the context window, we significantly expand the page size, resulting in stronger compression (\Cref{subsec:gist_mem}).
For example, the Gutenburg gists from the test set have 2,217 words on average and the maximum is 6,471 words, whereas the movie script gists have 2,155 words on average and the maximum is 4,511 words.
\Cref{fig:narrativeqa_gutenberg_stats,fig:narrativeqa_movie_stats} show the word statistics for the original text and the gists in Gutenberg and movie scripts respectively.
The compression rate of the gists is 96.80\% for Gutenberg texts and 91.98\% for movie scripts.
See \Cref{appendix:details,appendix:additional_narrativeqa} for NarrativeQA pagination hyperparameters and more details.

\begin{table*}[t!]
    \setlength{\tabcolsep}{0.2em}
    \footnotesize  %
    \centering
    \begin{tabular}{l|c|c|c|c|c|c|c}
        \hline
        Method &    CR (\# LU) & LLM Rating-1 & LLM Rating-2 & ROUGE-1 & ROUGE-2 & ROUGE-L & Resp. Length\\
        \hline
        \multicolumn{7}{l}{BM25 Retrieval} \\
        \hline
        \hspace{3mm} Top-1 &    95.69\% (1.00) & 32.48\% $\pm$ 1.65 & 63.85\% $\pm$ 1.51 & 27.53 $\pm$ 0.23 & 7.00 $\pm$ 0.14 & 18.45 $\pm$ 0.16 & 48.62 $\pm$ 0.28 \\
        \hspace{3mm} Top-2 &    91.48\% (2.00) & 29.41\% $\pm$ 0.60 & 71.57\% $\pm$ 1.48 & 28.85 $\pm$ 0.17 & 7.59 $\pm$ 0.08 & 19.34 $\pm$ 0.14 & 52.39 $\pm$ 0.49 \\
        \hspace{3mm} Top-3 &    86.93\% (3.00) & 34.80\% $\pm$ 1.14 & 79.53\% $\pm$ 0.35 & 30.69 $\pm$ 0.17 & 8.40 $\pm$ 0.11 & 20.64 $\pm$ 0.13 & 53.59 $\pm$ 0.35 \\
        \hspace{3mm} Top-4 &    82.55\% (4.00) & 35.66\% $\pm$ 0.30 & 81.13\% $\pm$ 0.35 & 31.10 $\pm$ 0.10 & 8.53 $\pm$ 0.06 & 20.36 $\pm$ 0.11 & 54.96 $\pm$ 0.42 \\
        \hspace{3mm} Top-5 &    78.13\% (5.00) & 39.09\% $\pm$ 0.92 & 84.44\% $\pm$ 0.46 & 31.16 $\pm$ 0.14 & 8.52 $\pm$ 0.08 & 20.69 $\pm$ 0.03 & 54.52 $\pm$ 0.13 \\
        \hspace{3mm} Top-6 &    73.97\% (6.00) & 37.87\% $\pm$ 0.90 & 83.70\% $\pm$ 0.87 & 31.06 $\pm$ 0.04 & 8.38 $\pm$ 0.06 & 20.43 $\pm$ 0.08 & 56.18 $\pm$ 0.44 \\
        \hline
        \multicolumn{7}{l}{Neural Retrieval with Gemini API} \\
        \hline
        \hspace{3mm} Top-1 &    95.99\% (1.00) & 34.80\% $\pm$ 1.39 & 68.87\% $\pm$ 0.62 & 27.86 $\pm$ 0.12 & 7.12 $\pm$ 0.04 & 18.76 $\pm$ 0.09 & 49.46 $\pm$ 0.23 \\
        \hspace{3mm} Top-2 &    92.02\% (2.00) & 40.32\% $\pm$ 0.92 & 81.50\% $\pm$ 0.46 & 30.17 $\pm$ 0.08 & 8.03 $\pm$ 0.03 & 19.80 $\pm$ 0.08 & 55.48 $\pm$ 0.27 \\
        \hspace{3mm} Top-3 &    87.93\% (3.00) & 40.93\% $\pm$ 1.35 & 85.17\% $\pm$ 1.25 & 31.36 $\pm$ 0.12 & 8.67 $\pm$ 0.10 & 20.68 $\pm$ 0.10 & 56.71 $\pm$ 0.27 \\
        \hspace{3mm} Top-4 &    83.71\% (4.00) & 40.56\% $\pm$ 0.62 & 84.31\% $\pm$ 0.87 & 31.52 $\pm$ 0.11 & 8.59 $\pm$ 0.10 & 20.40 $\pm$ 0.10 & 56.47 $\pm$ 0.71 \\
        \hspace{3mm} Top-5 &    79.47\% (5.00) & 40.20\% $\pm$ 0.76 & 86.76\% $\pm$ 0.60 & 31.32 $\pm$ 0.11 & 8.49 $\pm$ 0.11 & 20.49 $\pm$ 0.07 & 56.73 $\pm$ 0.91 \\
        \hspace{3mm} Top-6 &    75.44\% (6.00) & 40.81\% $\pm$ 0.52 & 87.01\% $\pm$ 0.35 & 31.92 $\pm$ 0.02 & 8.73 $\pm$ 0.09 & 20.82 $\pm$ 0.05 & 58.39 $\pm$ 0.31 \\
        \hline
        \multicolumn{7}{l}{Truncated Raw Content} \\
        \hline
        \hspace{3mm} First 6k words &    32.59\% (0.00) & 14.71\% $\pm$ 0.79 & 52.45\% $\pm$ 0.69 & 25.42 $\pm$ 0.05 & 4.98 $\pm$ 0.09 & 16.58 $\pm$ 0.10 & 58.42 $\pm$ 0.11 \\
        \hspace{3mm} Last 6k words  &    32.38\% (0.00) & 10.42\% $\pm$ 0.62 & 35.66\% $\pm$ 2.46 & 20.69 $\pm$ 0.19 & 3.44 $\pm$ 0.10 & 14.13 $\pm$ 0.08 & 44.23 $\pm$ 0.11 \\
        \hline
        \textbf{GistMem} &    83.13\% (0.00) & 40.20\% $\pm$ 0.96 & 89.83\% $\pm$ 0.76 & 31.00 $\pm$ 0.09 & 7.99 $\pm$ 0.04 & 20.15 $\pm$ 0.08 & 65.75 $\pm$ 0.20 \\
        \hline
        \multicolumn{7}{l}{{\textbf{ReadAgent-P}}} \\
        \hline
        \hspace{3mm} Look up 1 pg    &    80.00\% (0.98) & 40.56\% $\pm$ 0.46 & 89.46\% $\pm$ 1.48 & 31.26 $\pm$ 0.09 & 8.22 $\pm$ 0.15 & 20.29 $\pm$ 0.07 & 63.78 $\pm$ 1.13 \\
        \hspace{3mm} Look up 1-2 pgs &    77.38\% (1.71) & 39.71\% $\pm$ 1.87 & 89.71\% $\pm$ 0.60 & 31.11 $\pm$ 0.04 & 8.01 $\pm$ 0.15 & 20.21 $\pm$ 0.04 & 64.73 $\pm$ 1.02 \\
        \hspace{3mm} Look up 1-3 pgs &    75.07\% (2.53) & 38.36\% $\pm$ 1.21 & 89.71\% $\pm$ 0.60 & 31.50 $\pm$ 0.29 & 8.15 $\pm$ 0.15 & 20.45 $\pm$ 0.24 & 63.91 $\pm$ 1.58 \\
        \hspace{3mm} Look up 1-4 pgs &    73.48\% (3.08) & 39.95\% $\pm$ 1.51 & 90.56\% $\pm$ 0.35 & 31.34 $\pm$ 0.05 & 8.08 $\pm$ 0.18 & 20.26 $\pm$ 0.07 & 63.40 $\pm$ 0.79 \\
        \hspace{3mm} Look up 1-5 pgs &    72.29\% (3.50) & 37.99\% $\pm$ 0.96 & 87.75\% $\pm$ 0.46 & 31.16 $\pm$ 0.10 & 8.06 $\pm$ 0.05 & 20.35 $\pm$ 0.12 & 65.22 $\pm$ 1.40 \\
        \hspace{3mm} Look up 1-6 pgs &    70.90\% (3.97) & 39.09\% $\pm$ 2.04 & 88.24\% $\pm$ 0.60 & 31.50 $\pm$ 0.30 & 8.05 $\pm$ 0.13 & 20.26 $\pm$ 0.13 & 66.70 $\pm$ 0.62 \\
        \hline
        \textbf{ReadAgent-S} 1-6 pgs &    70.34\% (3.55) & \textbf{46.57\%} $\pm$ 0.87 & \textbf{91.54\%} $\pm$ 0.30 & \textbf{32.90} $\pm$ 0.17 & \textbf{8.87} $\pm$ 0.23 & \textbf{21.15} $\pm$ 0.14 & 68.87 $\pm$ 0.60 \\
        \hline
    \end{tabular}
    \caption{
        \textbf{QMSum validation} results (PaLM 2-L) means and standard deviations across 3 runs.
        35 articles and 272 questions.
        \textbf{CR} is the compression rate.
        \textbf{\# LU} is the number of lookups.
        \textbf{Resp. Length} is the length in words of the model's final response.
    }
    \label{tab:qmsum_validation}
\end{table*}

For the neural retrieval models, we use the gist memory embedding vectors rather than the page embedding vectors because the Gemini API embedding model is limited to 10,000 characters (or less than 2,000 tokens, in expectation), which is too short for embedding full pages in our NarrativeQA experiments.
However, using those embedding vectors, we then return the original pages to the LLM context as normal, and use those pages as described in \Cref{subsec:baseline}.

Because the Gutenberg texts and the movie scripts have significantly different distributions, we present the results separately in \Cref{tab:narrativeqa_results}.
ReadAgent again outperforms all the baselines across all subsets of NarrativeQA.

\subsubsection{QMSum}
\label{sec:qmsum}

\begin{figure}[htb]
    \centering
    \includegraphics[width=0.45\textwidth]{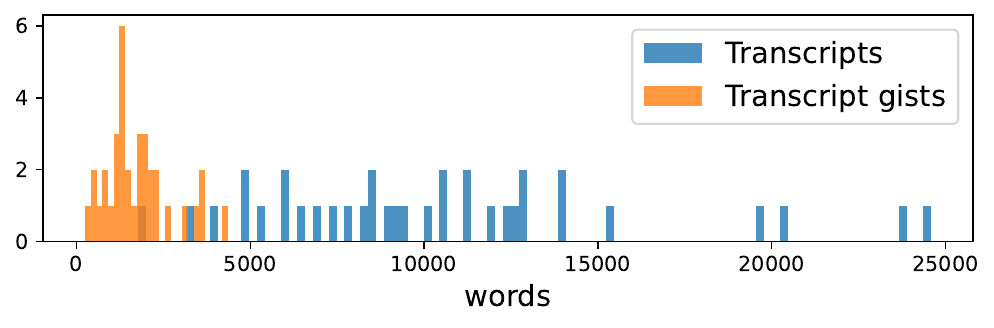}
    \vspace{-0.3cm}
    \caption{
        Histogram of QMSum word counts for the original transcripts and the gisted transcripts.
        The gisted transcripts are all less than 5,000 words, allowing them to entirely fit into the context window of PaLM 2-L.
    }
    \label{fig:qmsum_val_hist}
\end{figure}

QMSum~\citep{zhong2021qmsum} consists of meeting transcripts on various topics and associated questions or instructions.
We use the concatenated version of QMSum provided by SCROLLS~\citep{shaham2022scrolls}.
The transcripts tend to be quite long, ranging in length from 1,000 to 26,300 words, with an average length of about 10,000 words.
\Cref{fig:qmsum_val_hist} shows the histograms of word counts for the QMSum training set.
The answers are free form text, so the standard evaluation metric is ROUGE F-Measure.
We additionally evaluate using our LLM Ratings (\Cref{subsec:llm_rater}).
See \Cref{appendix:details,appendix:additional_qmsum} for hyperparameters and additional results.

In \Cref{tab:qmsum_validation} and \Cref{tab:qmsum_test} (Appendix), we see that performance improves as the compression rate decreases, so techniques that look up more pages tend to do better than techniques that look up fewer pages.
We also see that ReadAgent-S substantially outperforms ReadAgent-P (and all baselines).
This performance improvement comes at a cost of up to six times as many requests in the retrieval phase.
Since other datasets don't have such a strong performance improvement, we suspect that QMSum is in some sense a more challenging dataset, requiring the model to actively search through the gisted transcript to locate relevant information.
This hypothesis seems reasonable, as meeting transcripts are much less structured than the documents, books, and movies found in QuALITY and NarrativeQA.

A large fraction of the tasks in QMSum are a request to provide a summary, rather than a concrete question about some content in the meeting.
For many of these, the LLM refuses to look up any pages, instead responding with ``I don't need to look up any pages. I can summarize the whole meeting based on what I already remember.'', for example.
Consequently, the average number of pages looked up for ReadAgent is much lower than the maximum allowed.
However, on the tasks that actually involve a question, ReadAgent tends to use most or all of the available lookup pages.

In \Cref{tab:qmsum_validation,tab:qmsum_test}, the ROUGE scores by themselves don't always show a clear trend.
This is because as the length of the texts increase (corresponding to the compression rates decreasing), the response lengths increase as well.
Longer response lengths result in lower ROUGE precision values, which pushes down the F-Measures.
Consequently, for the ROUGE scores to increase as text length increases, the improvement to recall must be more substantial than the reduction to precision.
This happens to some extent, but the effect size is small.
Furthermore, including gists in the text substantially increases the response length, as is the case for GistMem and all the ReadAgent approaches.
This increase is in spite of the fact that all models use the same question-answering prompt, so there is no prompt difference to cause the increased response lengths.
This makes it much more challenging for GistMem and ReadAgent to outperform the retrieval methods in ROUGE score.
Nevertheless, ReadAgent-S manages to have the highest ROUGE scores as well as the highest LLM ratings.
Because of these issues with ROUGE, we consider the LLM ratings to be more informative for comparisons between these runs.
However, the LLM ratings do not make it easy to compare with results using a different LLM to rate, such as GPT, and they also do not allow for easy comparisons with other works.
The same observation applies to the NarrativeQA results above.

\subsection{Ablation Study and Analysis}
\label{subsec:ablation}

\paragraph{Retrieval Quality}
In \Cref{tab:retrieval_comparison}, we compare using GistMem with neural retrieval to look up one page with using ReadAgent to look up one page.
This is equivalent to replacing ReadAgent's prompt-based retrieval with neural retrieval.
ReadAgent's retrieval performs better here.

\begin{table}[htb]
    \small
    \centering
    \begin{tabular}{l|c}
        \hline
        Method & Accuracy \\
        \hline
        GistMem + Neural Retrieval Top-1 & 82.65\% \\
        ReadAgent-P (Look up 1 pg) & 84.13\% \\
        \hline
    \end{tabular}
    \caption{%
        ReadAgent retrieval vs. GistMem with neural retrieval.
    }
    \label{tab:retrieval_comparison}
    \vspace{-3mm}
\end{table}

\paragraph{Episode pagination}
In this work we ask ReadAgent to decide where to pause reading and what information to store together in memory (\Cref{subsec:gist_mem}), whereas in prior art, rule-based segmentation of text is typically used~\cite{chen2023walking,wu2021recursively}.
We compare the two approaches with similar page length on average in \Cref{tab:pagination_ablation} to demonstrate that it is indeed beneficial to break at pause points that LLMs consider natural (e.g. scene transitions, ends of dialogue, narrative transitions, etc).

\begin{table}[h!]
    \small
    \centering
    \begin{tabular}{l|c|c}
        \hline
         & LLM & Uniform Length \\
        \hline
        ReadAgent-P (1-5 pgs) Acc. & \bf{86.83\%} & 85.71\% \\
        \hline
    \end{tabular}
    \caption{%
        ReadAgent accuracy on QuALITY with episode pagination based on LLM (PaLM 2-L) vs. uniform length pagination.
    }
    \label{tab:pagination_ablation}
    \vspace{-3mm}
\end{table}

\paragraph{The compression trade-off}
\Cref{tab:compression_trade_off} presents the empirical results of compression rate increasing as page size increases.
As the compression rate decreases, the gists are more useful for answering questions directly.
However, for ReadAgent with look-ups, when the initial gist compression rate gets too high, accuracy suffers.

\begin{table}[h!]
    \small
    \centering
    \begin{tabular}{l|c|c|c|c}
        \hline
         & \multicolumn{2}{c}{GistMem} & \multicolumn{2}{c}{ReadAgent-P (1-5 pgs)} \\
        \texttt{max\_words} & CR & Acc & CR & Acc \\
        \hline
        400 &  81.81\% & 78.91\% & 66.71\% & 86.82\% \\
        600 &  85.53\% & 77.52\% & 66.45\% & 86.83\% \\
        800 & 88.12\% & 76.22\% & 65.06\% & 86.34\% \\
        1200 & 91.38\% & 73.97\% & 61.77\% & 85.67\% \\
        \hline
    \end{tabular}
    \caption{%
        Compression rate increases as the maximum number of words allowed per page increases on QuALITY.
        Our default setting of min/max words is 280/600.
        In the other three experiments, we scale min words proportionally with max words.
    }
    \label{tab:compression_trade_off}
    \vspace{-3mm}
\end{table}

%% file: appendix.tex
\section{Author Contributions}
\label{appendix:contribution}

\textbf{Kuang-Huei Lee} developed the initial working prototype, the method and the experiments on QuALITY and NarrativeQA, was a main writer of the manuscript, and led the project overall.

\textbf{Xinyun Chen} developed the method, the LLM rater, and experiments on NarrativeQA, and significantly contributed to manuscript writing.

\textbf{Hiroki Furuta} developed the web navigation experiments, and significantly contributed to manuscript writing.

\textbf{John Canny} contributed in the initial conceptualization, advised the project, and helped with manuscript editing.

\textbf{Ian Fischer} co-proposed the core idea, developed the method and experiments on QMSum, and was a main writer of the manuscript.

\section{Evaluation with GPT-3.5}
\label{appendix:gpt}

\Cref{tab:quality_results_gpt} shows the results of running experiments using exactly the same setup as described in \Cref{sec:quality}, but using GPT 3.5 Turbo rather than PaLM 2-L.
GPT 3.5 Turbo has a context length of over 16,000 tokens, so the QuALITY dataset easily fits into context.
We don't specifically tune prompts for GPT 3.5 Turbo, but instead use the same prompts that we use for PaLM 2-L.
GPT 3.5 Turbo has a much harder time with this task than PaLM 2-L, but the same general trends hold.
Neural Retrieval is weaker than ReadAgent.
ReadAgent-S achieves comparable performance to using the full article content.
The gap between ReadAgent-P and ReadAgent-S appears to be larger using this model, but we found that ReadAgent-P is very restrictive of how many pages to look up (1.0 in average) even though we allow up to 5.
We think that this can likely be remedied if we engineer the prompt for GPT 3.5 Turbo.
Nonetheless, comparing to using top 3 from neural retrieval, ReadAgent-P still yields better accuracy and compression rate.

\begin{table}[htb]
    \centering
    \begin{tabular}{l|c|c}
        \hline
        Method & CR (\# LU) & Accuracy \\
        \hline
        Neural Retrieval with Gemini API Top-3 & 73.13\% (3) & 69.22\% \\
        Full Raw Content & 0\% & 73.30\% \\
        \textbf{GistMem} & 84.24\% & 66.06\% \\
        \textbf{ReadAgent-P} Look up 1-5 pgs & 76.60\% (1.0) & 69.65\% \\
        \textbf{ReadAgent-S} Look up 1-6 pgs & 60.43\% (3.4) & 72.10\% \\
        \hline
    \end{tabular}
    \caption{
    QuALITY results on the dev set of 230 docs and 2086 questions using GPT-3.5-turbo. 
    \textbf{CR} is the compression rate.
    \textbf{\# LU} is the number of lookups.
    We report 1 run for each experiment for cost considerations.
    } 
    \label{tab:quality_results_gpt}
\end{table}

\section{Pagination Hyperparameters}
\label{appendix:details}

\paragraph{Pagination Details}
As described in \Cref{subsec:gist_mem}, \texttt{max\textunderscore words} and \texttt{min\textunderscore words} are two episode pagination hyperparameters.
\Cref{tab:pagination_hparams} gives their values for each of the experiments in \Cref{sec:exp}.

{\renewcommand{\arraystretch}{1.5}%
\begin{table}[htb]
    \small
    \centering
    \begin{tabular}{c|c}
        Dataset & $\frac{\text{\texttt{max\textunderscore words}}}{\text{\texttt{min\textunderscore words}}}$  \\
        \hline
        QuALITY & $\frac{600}{280}$ \\
        QMSum   & $\frac{600}{280}$ \\
        NarrativeQA Gutenberg & $\frac{3000}{500}$ \\
        NarrativeQA movie scripts & $\frac{1000}{600}$
    \end{tabular}
    \caption{Pagination hyperparameters.}
    \label{tab:pagination_hparams}
\end{table}}

\clearpage
\section{Case Study}
\label{appendix:case_study}

In this section, we analyze reading comprehension examples to demonstrate where the ability to simultaneously think over long-range global context and focus on local information is important.
We selected the short story ``off course'' by Mack Reynolds\footnote{%
    Available at \url{http://aleph.gutenberg.org/3/0/0/3/30035//30035-h//30035-h.htm}.
}
because it is extremely short (2,712 words) and it is only broken into 8 pages, yet even so, neural retrieval using 4 pages gets three questions wrong that ReadAgent correctly answers.
For this story, ReadAgent answers 6 of 8 questions correctly.
Neural retrieval answers 3 of 8 correctly, and doesn't get either question correct that ReadAgent misses.
Note that in all three examples, ReadAgent only chooses to select two pages, even though it is also permitted to select up to 4.
This flexibility is another advantage that ReadAgent has over standard retrieval systems.

\begin{tcolorbox}[title=``off course'' Gist Memory]
\small
$\langle \text{P}0 \rangle$ Patrolmen Dermott and Casey encounter Dameri Tass, an alien who has landed on Earth. Dameri attempts to communicate with them using a device that translates his thoughts into English.

$\langle \text{P}1 \rangle$ The alien Dameri Tass used a helmet to learn English from Tim Casey, an Irish patrolman. He then became fascinated by a horse and wanted to use the helmet on the animal. Patrolman Dermott felt like he was in a shaggy dog story.

$\langle \text{P}2 \rangle$ A helicopter arrived, interrupting the horse's inspection. Two Army officers exited and ordered a police cordon around the spacecraft. The alien spoke, surprising the general. More police and military personnel arrived.

$\langle \text{P}3 \rangle$ Dameri Tass, an alien visitor, was whisked away to Washington and held incommunicado for several days. His arrival caused a global furor. Officials worried about the potential impact of his message on society. Eventually, the UN demanded that he be allowed to speak before the Assembly. The White House agreed and a date was set.

$\langle \text{P}4 \rangle$ The world eagerly awaited a message from space. Dameri Tass, an envoy from a super-civilization, was expected to guide the world. Most people were ready to be guided, but some were not. The U.N. Secretary-General was nervous about introducing the envoy, as they knew very little about him. He had been asleep for most of his time on Earth and had only recently woken up. He spent his time playing with a dog, cat, and mouse. The Secretary-General was worried about what the envoy would say.

$\langle \text{P}5 \rangle$ Dameri Tass, an alien, is brought to Earth and mistaken for an envoy from another planet. He reveals he is just a collector for a zoo.

$\langle \text{P}6 \rangle$ Dameri Tass, an alien, mistakenly landed on Earth. He addressed a large crowd, criticizing their weapons, wars, and lack of a planet-wide government. He then left, refusing to take any Earth creatures with him, but expressing interest in horses.

$\langle \text{P}7 \rangle$ The others watched as the first visitor from space hurriedly left Earth.
\end{tcolorbox}

\begin{wraptable}[7]{R}{0cm}
    \small
    \centering
    \begin{tabular}{c|l}
        Page \# & Starting sentence in text \\
        \hline
        0 & Shure and begorra... \\
        1 & The alien stooped down... \\
        2 & Interest in the horse was ended... \\
        3 & ``Sure, and it's quite a reception''... \\
        4 & Excitement, anticipation... \\
        5 & ``Here he comes,''... \\
        6 & He straightened and started off... \\
        7 & The others drew back...
    \end{tabular}
    \vspace{-0.35cm}
    \caption{Pagination for ``off course''.}
    \label{tab:off_course_pagination}
\end{wraptable}

\paragraph{Distracting retrieval}
The first question gives an example of retrieval of distracting pages and the lack of global context provided by the gist memory causing the LLM to select the incorrect answer when using neural retrieval, even though it had also retrieved the pages that should have led to the correct answer.
We provide the gist memory below and the story's pagination in \Cref{tab:off_course_pagination}.

\begin{tcolorbox}[title=``off course'' Question 1]
\small
What was Dameri’s purpose in landing on earth?

(A) He wanted to witness an uncivilized planet and share knowledge

(B) His spaceship needed to land for repairs

(C) He heard reports that Earth had interesting animal specimens for his collection

(D) He arrived on accident while exploring planets in the Galactic League

The correct answer is (D).
ReadAgent chose (D).
Neural retrieval chose (C).
\end{tcolorbox}

For the question above, ReadAgent looked up pages 5 and 6.
Neural retrieval looked up pages 3, 4, 5, and 6.
Pages 4 and 5 both make prominent mention of animals, and Page 5 explicitly mentions that the alien is a collector for a zoo, so answer (C) seems reasonable based on the information on those pages.
However, Pages 5 and 6, together with the global context from the gist memory, make it clear that (D) is the correct answer.
Since neural retrieval provided both of those pages, the lack of the global context combined with the additional distractor pages led the LLM astray.

\begin{tcolorbox}[title=``off course'' Question 2]
\small
What happened to Dameri while he was in custody of the government?

(A) He picked up an accent from the guards

(B) He slept almost the entire time

(C) He learned horses were creatures that could be ridden

(D) He was too shy to speak

The correct answer is (B).
ReadAgent chose (B).
Neural retrieval chose (A).
\end{tcolorbox}

\paragraph{Incorrect retrieval}
The same story provides two examples of the consequences of incorrect retrieval, and the benefits of the gist memory.
For the question above, ReadAgent looked up pages 3 and 4.
Neural retrieval looked up pages 0, 1, 3, and 6.
The correct answer is clearly stated on Page 4, and also clearly stated in the gist of Page 4.
If the LLM had access to either of those, it should have been able to answer correctly.
Instead, it was undoubtedly confused by Pages 0 and 1, where the alien learns an accent from one of the police officers in the initial encounter.

\begin{tcolorbox}[title=``off course'' Question 3]
\small
How did Dameri Tass communicate in English?

(A) He could communicate telepathically

(B) He never was able to communicate in English

(C) He used a handheld translation device

(D) He acquired the knowledge from a human

The correct answer is (D).
ReadAgent chose (D).
Neural retrieval chose (C).
\end{tcolorbox}

For the question above, ReadAgent looked up pages 0 and 1.
Neural retrieval looked up pages 0, 3, 4, and 6.
The critical information was in Page 1, although Page 0 was also relevant.
The remaining pages were only relevant in that they demonstrated that (B) was incorrect.
Again, the gist memory was sufficient to answer the question correctly, in addition to providing clear signal about what pages are relevant to the question.
But neural retrieval's selection of Page 0 without Page 1 made (C) seem plausible, as Page 0 discusses a device that the alien was clearly trying to use for communication.

\section{ReadAgent for Web Navigation}
\label{appendix:web_agent}

We made an attempt to extend ReadAgent to decision making tasks.
In particular, we apply ReadAgent for autonomous \textit{web navigation}~\citep{shi2017miniwob,kim2023rci,furuta2024multimodal}, where the goal is to autonomously control browsers or computer interfaces to complete tasks with natural language instructions provided by users.
Such instruction would be something like \textit{Book an appointment for applying new passport for one adult, Ellen Walker, with phone number 123-456-7890 and email address EW@gmail.com on April 4, 2023 at 1 pm in the post office nearest to zip code 60505. Don't send updates via text message}).
Example web agent actions include \textit{click}, \textit{type}, and \textit{select} (e.g. \textit{click}, \textit{type nearest post office}, \textit{select April 4, 2023}).
Because real-world websites can have very long HTML, LLM web agents often struggle with context length if it operates on raw content~\citep{gur2023realworld}.

\subsection{Implementation}

\paragraph{Pagination}
For HTML, we leverage the explicit HTML DOM tree structure, decomposing the HTML into snippets with elements at a target depth and their descendants. %
We test the depth from 5 to 7 and choose the best.
We use these snippets as the ``pages'' instead of asking the LLM to paginate.

\paragraph{Memory Gisting}
Similar to ReadAgent for reading comprehension, we prompt the LLM to summarize snippets into gists zero-shot, and subsequently concatenate the gists.
We contextualize the gists with snippet index number in a python dictionary-format (e.g. \{``index'': ..., ``content'': ...\}).

\paragraph{Interactive Look-up}
In the interactive look-up step, the LLM looks at a given task instruction, previous action history, and the gists to decide which original HTML snippets it wants to look up.
We experimented with parallel look-up (ReadAgent-P) in the web navigation setting for faster experiments.
Finally, to predict next-step actions, the LLM reads the retrieved snippets again and predicts the target element id to interact with, the type of action operation (click, type or select), and the input value (if any).

\subsection{Mind2Web}
We evaluate ReadAgent for Web Navigation on the Mind2Web~\citep{deng2023mind2web} dataset, a real-world planning and web action prediction benchmark, consisting of 2K instructions and episodes collected from 137 websites.
The agent's task is to predict the next-step action (click, type and select) given HTML, task instruction, and previous action history.
Mind2Web has three test set splits: cross-task (252 tasks from 69 websites), cross-website (177 tasks from 10 websites), and cross-domain (912 tasks from 73 websites), which was originally designed for different testing different type of generalization.
However, since our approach is zero-shot without training, these splits do not serve their original purposes.

\paragraph{Baselines}
MindAct from the Mind2Web paper~\citep{deng2023mind2web} first uses a DeBERTa-base~\citep{he2020deberta} model trained for task-relevant element retrieval to get the top 50 relevant elements.
Instead of directly predicting target element id (part of an action), it formulates this task as iterative multi-choice question-answering with target element ids sampled from the top 50 and uses the LLM to solve it for performance purpose (see \citet{deng2023mind2web} for details).
The same LLM also predicts the type of action and an optional value.
MindAct (GPT-4) results are the state-of-the-art.
We additional generate MindAct results with PaLM 2-L as a reference.

Following the reading comprehension experiments~(\Cref{sec:exp}), we also compare with using full raw HTML, retrieval with BM25, neural retrieval with Gemini API embedding model (models/embedding-001), and using the gists without look-up, which, like ReadAgent, are not trained for web navigation tasks.
We ask the LLM to directly predict that target element id as it is a simpler and more tractable implementation in our setting.

\begin{table*}[htb]
\begin{center}
\begin{small}
\scalebox{0.65}{
\begin{tabular}{lrrrrrrrrrrrrrrrr}
\toprule
& \multicolumn{5}{c}{\texttt{Cross-Task}} & \multicolumn{5}{c}{\texttt{Cross-Website}} & \multicolumn{5}{c}{\texttt{Cross-Domain}} \\
 \cmidrule(r){2-6} \cmidrule(r){7-11} \cmidrule(r){12-16}
  & \textbf{CR} & \textbf{Ele. Acc} & \textbf{Op. F1} & \textbf{Step SR} & \textbf{SR} & \textbf{CR} & \textbf{Ele. Acc} & \textbf{Op. F1} & \textbf{Step SR} & \textbf{SR} & \textbf{CR} & \textbf{Ele. Acc} & \textbf{Op. F1} & \textbf{Step SR} & \textbf{SR} \\
\midrule
\multicolumn{5}{l}{\textbf{Using supervisedly trained RankLM}} \\
\midrule
\textbf{MindAct} (GPT-3.5 + Rank LM$^{*}$) & -- & 20.3 & 56.6 & 17.4 & 0.8 & -- & 19.3 & 48.8 & 16.2 & 0.6 & -- & 21.6 & 52.8 & 18.6 & 1.0\\
\textbf{MindAct} (GPT-4 + Rank LM$^{*}$) & -- & \textbf{41.6} & 60.6 & \textbf{36.2} & 2.0 & -- & 35.8 & 51.1 & 30.1 & 2.0 & -- & 37.1 & 46.5 & 26.4 & 2.0\\
\textbf{MindAct} (PaLM 2-L + Rank LM$^{*}$) & -- & 29.8 & 61.9 & 24.4 & 1.2 & -- & 28.8 & 59.6 & 21.6 & 0.6 & -- & 29.9 & 60.4 & 24.5 & 1.3\\
\midrule
\midrule
\multicolumn{5}{l}{\textbf{No training}} \\
\midrule
(PaLM 2-L) \\
~~+\textbf{Raw HTML} & 0.0 & 22.1 & \underline{\textbf{76.7}} & 19.2 & 1.2 & 0.0 & 22.2 & 72.3 & 18.2 & 1.7 & 0.0 & 23.6 & 75.6 & 20.9 & 1.0\\
~~+\textbf{BM25 Retrieval} (Top-1) & 43.7 & 16.3 & 61.7 & 14.2 & 0.4 & 49.7 & 17.8 & 60.8 & 15.2 & 0.0 & 51.6 & 17.3 & 60.4 & 15.9 & 0.0 \\
~~+\textbf{BM25 Retrieval} (Top-5) & 19.5 & 25.9 & 70.4 & 22.4 & 2.0 & 17.6 & 29.5 & 71.8 & 23.1 & 1.7 & 19.2 & 27.6 & 71.1 & 24.4 & 1.0 \\
~~+\textbf{Neural Retrieval} (Top-1) & 74.4 & 14.6 & 55.5 & 11.7 & 0.4 & 87.9 & 18.0 & 55.8 & 14.0 & 0.0 & 82.8 & 16.4 & 60.3 & 14.2 & 0.1\\
~~+\textbf{Neural Retrieval} (Top-5) & 32.4 & 26.4 & 71.9 & 22.6 & 0.8 & 37.2 & 26.7 & 69.1 & 22.3 & 2.8 & 38.1 & 30.0 & 72.5 & 26.9 & 1.2\\
\midrule
\textbf{GistMem} & 84.4 & 11.7 & 43.1 & 9.5 & 0.0 & 82.5 & 11.7 & 43.6 & 8.4 & 0.0 & 83.0 & 13.4 & 49.6 & 11.7 & 0.5 \\
\textbf{ReadAgent-P: Lookup 1 snippet} & 55.1 & 31.1 & 70.1 & 26.8 & 2.0 & 54.1 & 34.5 & 74.1 & 28.2 & 2.3 & 55.2 & 36.1 & 75.6 & 33.0 & 2.0\\
    \textbf{ReadAgent-P: Lookup 1-5 snippets} & 35.9 & \underline{33.7} & 72.5 & \underline{29.2} & \underline{\textbf{2.8}} & 35.6 & \underline{\textbf{37.4}} & \underline{\textbf{75.1}} & \underline{\textbf{31.1}} & \underline{\textbf{3.4}} & 48.2 & \underline{\textbf{37.2}} & \underline{\textbf{76.3}} & \underline{\textbf{33.4}} & \underline{\textbf{2.3}} \\
\midrule
\midrule
$\Delta(\text{Raw}-\text{ReadAgent})$ & -- &  \textcolor{cb_green}{+11.6} & \textcolor{cb_red}{-4.2} & \textcolor{cb_green}{+10.0} & \textcolor{cb_green}{+1.6} & -- &  \textcolor{cb_green}{+15.2} & \textcolor{cb_green}{+2.8} & \textcolor{cb_green}{+12.9} & \textcolor{cb_green}{+1.7} & -- &  \textcolor{cb_green}{+13.6} & \textcolor{cb_green}{+0.7} & \textcolor{cb_green}{+12.5} & \textcolor{cb_green}{+1.3} \\
$\Delta(\text{MindAct}-\text{ReadAgent})$ & -- & \textcolor{cb_green}{+3.9} & \textcolor{cb_green}{+10.6} & \textcolor{cb_green}{+4.8} & \textcolor{cb_green}{+1.6} & -- & \textcolor{cb_green}{+8.6} & \textcolor{cb_green}{+15.5} & \textcolor{cb_green}{+9.5} & \textcolor{cb_green}{+2.8} & -- & \textcolor{cb_green}{+7.3} & \textcolor{cb_green}{+15.9} & \textcolor{cb_green}{+8.9} & \textcolor{cb_green}{+1.0} \\
\bottomrule
\end{tabular}
}
\end{small}
\end{center}
\vskip -0.1in
\caption{
Web navigation performance on Mind2Web~\cite{deng2023mind2web}.
$^{*}$ marks models that are trained supervisedly for the web domain.
GistMem and ReadAgent results are all also based on PaLM 2-L.
We evaluate the performance in element accuracy (Ele. Acc), operation F1 (Op. F1), step success rate (Step SR), and episode success rate (SR).
We also measure the compression rate (CR).
The best performance across all the baselines is \textbf{bolded}, and the best across the approaches using PaLM 2-L is \underline{underlined}.
ReadAgent achieves consistently better performance than using raw HTML inputs (PaLM 2-L), retrieval methods, and MindAct (PaLM 2-L) with a trained Rank LM for HTML snippet retrieval.
}
\label{tab:mind2web}
\end{table*}

\subsection{Results}
As shown in \Cref{tab:mind2web}, ReadAgent achieves strong performance compared to the baselines.
In particular, the results are even better than MindAct (PaLM 2-L), which uses the supervisedly learned Rank LM, despite ReadAgent not using models trained on the web navigation domain.
Prior work shows that state-of-the-art LLMs alone are generally still weaker than the approaches using models specifically trained for the web navigation domain~\citep{furuta2023language}.

\begin{figure*}[htb]
\begin{minipage}[c]{0.65\textwidth}
    \centering
    \includegraphics[width=0.7\textwidth]{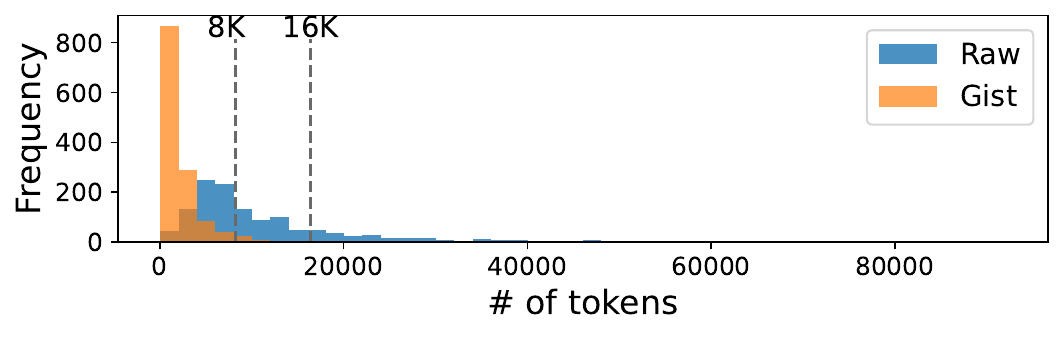}
\end{minipage}
\begin{minipage}[c]{0.35\textwidth}
\begin{center}
\begin{small}
\scalebox{0.85}{
\begin{tabular}{rrr}
\toprule
\textbf{Threshold} & \textbf{Raw} & \textbf{Gist} \\
\midrule
4096 Tokens & 14.2\% & 88.6\% \\
8192 Tokens & 51.5\% & 97.4\% \\
16384 Tokens & 79.1\% & 100\%\\
\midrule
50th Percentile Tokens & 8018 & 989\\
90th Percentile Tokens & 25337 & 3596\\
95th Percentile Tokens & 35779 & 5741\\
99th Percentile Tokens & 55642 & 12569\\
\bottomrule
\end{tabular}
}
\end{small}
\end{center}
\end{minipage}
\caption{
    \textbf{(Left)} Histogram of raw HTML and gist tokens in the Mind2Web cross-website split.
    Most of the input gists require fewer than 8K tokens.
    \textbf{(Right)} Statistics of token counts of raw HTML and gists.
}
\label{fig:min2web_hist}
\end{figure*}

\Cref{fig:min2web_hist} shows that gisting effectively reduces the number input tokens.
Most of the input gists require less than 8K tokens.
For example, 97.4\% of gisted inputs in cross-website split fits into the 8k context length, while only 51.5\% of raw HTML can fit in the context window.
The inputs are truncated for the parts that exceed the context length limit, which can significantly impact performance.

The results in \Cref{fig:min2web_hist,tab:mind2web} indicate that even using the gist memory and ReadAgent retrieval causes truncation on many web pages.
This is because the retrieved snippets are quite large, causing the compression rate to drop substantially.
In spite of those issues, the ReadAgent results give real gains over using the full context.
This indicates that even the truncated gists and retrieved pages are more informative than the truncated raw HTML when using an LLM with a small context length.

\section{Prompt Design}
\label{appendix:prompt_design}

We discuss our prompt design in this section to clarify some of our design decisions.
In most cases, the exact phrasing of the prompt has negligible impacts on the outcome.
For example, saying "You don’t need to answer the question yet." versus "DO NOT answer the question in your response." in the look-up prompts did not lead to significant differences in the results and they can be used interchangeably.

The exact prompts that we use for each dataset can be found on \url{https://read-agent.github.io/}.

\paragraph{Pagination Prompt}
In the QuALITY and QMSum experiments where the target page length is short and the \texttt{max\_words} threshold is low, we tried including the previous page in the pagination prompt (\Cref{subsec:gist_mem}), as we thought it could be helpful to have more surrounding context.
However, it appears that it only benefits our QMSum experiments, compared to not including the previous page.
We will leave this to future studies.

\paragraph{Look-up Prompt}
On QuALITY, we found adding "Take a deep breath and tell me: Which page(s) would you like to read again?" at the end of the look-up prompt improves response quality, when using PaLM 2-L and GPT-3.5.  
This is similar to what \citet{yang2023large} found.

\paragraph{Response Prompt}
For tasks that have free-form answers (QMSum and NarrativeQA), we explicitly ask the model to provide short and concise answers as follows. 
This helps mitigate the problem of verbose answers hurting ROUGE scores as we noted in \Cref{sec:qmsum}.

\begin{tcolorbox}[title=Example Response Prompt for Free-form Answers]
\small
\{GISTS AND EXPANDED PAGES\}
\vspace{1mm}

Question:
\{QUESTION\}
\vspace{1mm}

Answer the question based on the above passage and retrieved pages. Your answer should be short and concise.
\end{tcolorbox}

For multiple-choice questions, we query with the following prompt and parse the response to extract the model choice.

\begin{tcolorbox}[title=Example Response Prompt for Multiple-Choice Questions]
\small
Read the following article and answer a multiple choice question.
For example, if (C) is correct, answer with ``Answer: (C) ...``
\vspace{1mm}

Article:
\{GISTS AND EXPANDED PAGES\}
\vspace{1mm}

Question:
\{QUESTION\}
\{OPTIONS\}
\end{tcolorbox}

\paragraph{LLM Rater Prompts}
In \Cref{subsec:llm_rater}, we introduce Strict LLM Rater Prompt for exact matches and Permissive LLM Rater Prompt for partial matches.
The main reason for this design is that measuring exact matches can be overly strict as it disqualifies model answers with additional details. 
For example, assuming the reference answer is “the British” and the model answer is “the British army” (correct but with more details), using LLMs with “Strict LLM Rater Prompt” will tell us that this is not an exact match.
Empirically, we found these LLM ratings aligning well with our own judgements.

When referring to model's responses in the LLM rater prompts, we choose to refer them with ``John’s answer to the question was:`` because we want to prompt the LLM to judge whether the model-generated answer matches with the reference answer from a more objective, third-person view. 
Prior work, such as \cite{zheng2023judging}, uses “Assistant” to refer to responses from different model calls.
Here we instead choose a common name (John) that appears to be more natural.

\section{ReadAgent Variants}
\label{appendix:variants}

\subsection{Unconditional and Conditional ReadAgent}
\label{sec:uncond_cond_readagent}

When working with a long text, it is possible that the user will know ahead of time what task is to be solved.
In that case, conceivably the gisting step could include the task description in the prompt.
In so doing, it is easy to imagine that the LLM could do a better job of compressing out information that is irrelevant to the task, thereby improving efficiency and reducing distraction.
This approach would be \emph{Conditional ReadAgent}.
However, more generally, the task may not be known while preparing the gists, or it may be known that the gists need to be used for multiple different tasks, such as answering many questions about the text.
Thus, by excluding the task in the gisting step, the LLM may produce more broadly useful gists, at the cost of reduced compression and increased distracting information.
This setting would be \emph{Unconditional ReadAgent}.
We only explore the unconditional setting in this work, but we note that the conditional setting may be preferred in some situations.

\subsection{ReadAgent for Specific Domains}

Related to \Cref{sec:uncond_cond_readagent}, when applying ReadAgent to specific domains, it might be helpful to provide domain-specific instructions.
For example, if we apply ReadAgent to understanding a programming library, it could be useful to provide more specific instruction to LLMs to extract abstract descriptions of things like purpose of the code, functionalities, important signatures of functions or classes from each file as gists.

\subsection{Iterative Gisting}

For a very long event history, such as a conversation, we may consider further compression of the older memory with iterative gisting to allow having longer contexts, similar to older memories of humans being fuzzier.
Though this is not in the scope of this work, it may be useful for applications such as assistant agents, where context lengths can grow arbitrarily long over time as the user interacts with the agent.

\section{Comparing ReadAgent and MemWalker}
\label{appendix:memwalker}

As discussed in \Cref{sec:related_work}, similar to our work, MemWalker~\cite{chen2023walking} also reads long documents interactively like an agent through iterative prompting, instead of forcing LLMs to process everything at once.
It first constructs a summary tree where the lowest-level leafs are segments of raw text, the second level nodes are summaries of text segments, and the higher levels are summaries of summaries.
Given a task, it traverses the tree from the root to search for task-related information.
We think there are a few reasons to prefer the ReadAgent approach over MemWalker.

First, the reliability is a concern. 
Having LLMs traverse summary tree may not be a reliable process.
In our best-effort re-implementation of MemWalker with PaLM 2-L, it unsatisfyingly achieves 66.73\% on QuALITY.
To put that into perspective, using full raw content is 85.83\%, ReadAgent-P (look up 1-5 pages) is 86.63\%, ReadAgent-S (look up 1-6 pages) is 86.88\%, and using BM25 Top-1 is 70.55\%.
Part of the performance difference is caused by a high search failure rate.
11.7\% of the searches failed to finish after sufficient retries.
This failure rate of our implementation is in a similar range to what the authors reported: 91.4\% successes and 8.6\% failures\footnote{https://openreview.net/forum?id=H5XZLeXWPS}.
In contrast, the failure rate of ReadAgent is mostly 0\%. 

Second, the hierarchical summary structure makes it difficult to reason over related but distant information at the same granularity.
There isn't much detail preserved at the top levels of the hierarchy.
For example, if the two most important text pieces are at the beginning and the end of a very long text, the essential information could be in the first and last leaf. 
As the agent traverses down to the first leaf, it could be difficult to go back up to the root and down to the last leaf.

The motivations of the two approaches are also different.
MemWalker interacts with a summary tree and reasons over traversal trajectories, whereas ReadAgent interacts directly with documents and reasons over gist memories.

\section{NarrativeQA Additional Details}
\label{appendix:additional_narrativeqa}

\begin{figure}[h!]
    \centering
    \includegraphics[width=0.45\textwidth]{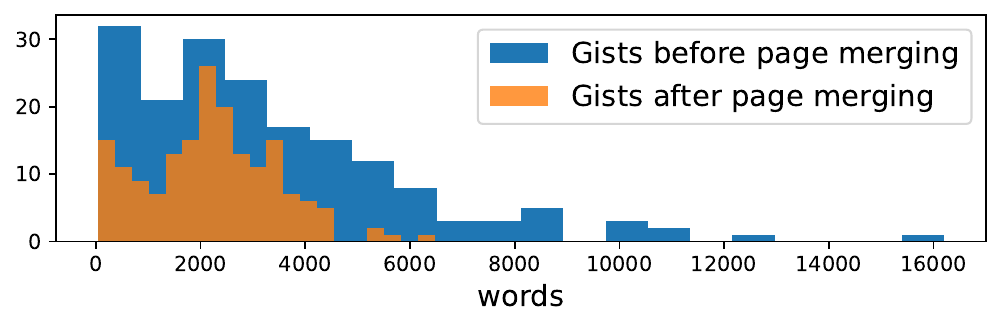}
    \caption{Histogram of NarrativeQA (Gutenberg) test set gists before and after page merging on the exceptionally long texts.}
    \label{fig:narrativeqa_gutenberg_stats_merging}
\end{figure}

\paragraph{Context Length Control}
As the NarrativeQA Gutenberg texts can be very long, the corresponding gists can sometimes exceed the context length.
For those exceptionally long texts, we ask the LLM to go through the pages and think whether it makes sense to merge pages iteratively with the following prompt, and then re-gist the new set of pages.
In so doing, we are able to increase the average page size and thus the compression rate (\Cref{fig:narrativeqa_gutenberg_stats_merging}).

\begin{tcolorbox}[title=Example NarrativeQA Gutenberg Page Merging Prompt]
\small
Given Page 1 and Page 2, please tell me whether Page 2 starts a new chapter/section/book that is different from what's in Page 1.
\vspace{1mm}

Please answer with yes, no, or not sure.
\vspace{1mm}

Page 1:

\{PREVIOUS PAGE TEXT\}
\vspace{1mm}

Page 2:

\{CURRENT PAGE TEXT\}

\end{tcolorbox}

The gists and pages can both be long for NarrativeQA. 
Thus, in the interactive look-up step of ReadAgent-P, we prevent the retrieved pages from exceeding the context length by asking the model to sort the pages by importance with the prompt below and iteratively detecting whether adding any pages could go beyond the context window.
For ReadAgent-S, we do a similar check to decide whether to early-stop the sequential look-up.

\begin{tcolorbox}[title=Example Parallel Lookup Prompt (ReadAgent-P) for NarrativeQA]
\small
The following text is what you remember from reading an article and a question related to it.
\vspace{1mm}

You may read 1, 2 or 3 page(s) of the article again to refresh your memory to prepare yourself for the question.
\vspace{1mm}

Please respond with which page(s) you would like to read in the order of importance, beginning with the most important page number.
\vspace{1mm}

For example, if you only need to read Page 8, respond with ``I want to look up Page [8] to ...''.
\vspace{1mm}

If you would like to read Page 12 and 7, respond with ``I want to look up Page [12, 7] to ...''.
\vspace{1mm}

If you would like to read Page 15, 2 and 3, respond with ``I want to look up Page [15, 2, 3] to ...''.
\vspace{1mm}

DO NOT select more pages if you don't need to.
\vspace{1mm}

You don't need to answer the question yet.
\vspace{1mm}

Text:

\{GIST MEMORY\}
\vspace{1mm}

Question:

\{QUESTION\}

\end{tcolorbox}

\section{Additional QMSum Results}
\label{appendix:additional_qmsum}

\begin{figure}[htb]
    \centering
    \includegraphics[width=0.45\textwidth]{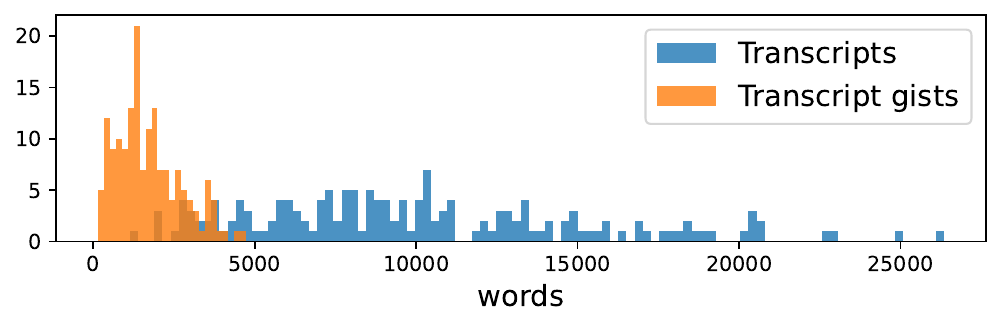}
    \caption{
        Histogram of QMSum word counts for the original transcripts and the gisted transcripts.
        The gisted transcripts are all less than 5,000 words, allowing them to entirely fit into the context windows of PaLM 2-L.
    }
    \vspace{-0.3cm}
    \label{fig:qmsum_hist}
\end{figure}

\Cref{fig:qmsum_hist} shows the histogram of word counts on the QMSum training set and the corresponding gist memories.

\Cref{tab:qmsum_test} shows the same results as \Cref{tab:qmsum_validation}, but on the QMSum test set.

\begin{table*}[b]
    \setlength{\tabcolsep}{0.2em}
    \footnotesize  %
    \centering
    \begin{tabular}{l|c|c|c|c|c|c|c}
        \hline
        Method &    CR (\# LU) & LLM Rating-1 & LLM Rating-2 & ROUGE-1 & ROUGE-2 & ROUGE-L & Resp. Length \\
        \hline
        \multicolumn{7}{l}{BM25 Retrieval} \\
        \hline
        \hspace{3mm} Top-1 &    95.61\% (1.00) & 24.67\% $\pm$ 0.44 & 66.90\% $\pm$ 0.87 & 28.81 $\pm$ 0.13 & 8.14 $\pm$ 0.15 & 19.62 $\pm$ 0.18 & 48.15 $\pm$ 0.18 \\
        \hspace{3mm} Top-2 &    91.32\% (2.00) & 31.79\% $\pm$ 1.31 & 79.95\% $\pm$ 0.67 & 30.89 $\pm$ 0.13 & 9.14 $\pm$ 0.05 & 20.67 $\pm$ 0.09 & 53.91 $\pm$ 0.64 \\
        \hspace{3mm} Top-3 &    87.25\% (3.00) & 33.45\% $\pm$ 0.00 & 83.63\% $\pm$ 1.05 & 31.39 $\pm$ 0.23 & 9.11 $\pm$ 0.05 & 21.03 $\pm$ 0.03 & 55.15 $\pm$ 0.61 \\
        \hspace{3mm} Top-4 &    82.86\% (4.00) & 37.72\% $\pm$ 1.05 & 86.12\% $\pm$ 0.50 & 31.71 $\pm$ 0.09 & 9.35 $\pm$ 0.13 & 21.26 $\pm$ 0.12 & 58.21 $\pm$ 0.37 \\
        \hspace{3mm} Top-5 &    78.79\% (5.00) & 39.38\% $\pm$ 1.02 & 86.60\% $\pm$ 0.44 & 32.66 $\pm$ 0.04 & \textit{\textbf{9.98}} $\pm$ 0.10 & \textbf{21.86} $\pm$ 0.05 & 59.20 $\pm$ 1.05 \\
        \hspace{3mm} Top-6 &    74.62\% (6.00) & 40.45\% $\pm$ 0.89 & 90.98\% $\pm$ 0.34 & 32.56 $\pm$ 0.03 & 9.78 $\pm$ 0.03 & 21.64 $\pm$ 0.09 & 60.40 $\pm$ 1.28 \\
        \hline
        \multicolumn{7}{l}{Neural Retrieval with Gemini API} \\
        \hline
        \hspace{3mm} Top-1 &    95.80\% (1.00) & 27.05\% $\pm$ 0.50 & 67.97\% $\pm$ 1.74 & 28.71 $\pm$ 0.12 & 7.98 $\pm$ 0.04 & 19.59 $\pm$ 0.04 & 49.76 $\pm$ 0.78 \\
        \hspace{3mm} Top-2 &    91.62\% (2.00) & 35.35\% $\pm$ 0.44 & 80.07\% $\pm$ 0.00 & 31.65 $\pm$ 0.18 & 9.59 $\pm$ 0.11 & 21.29 $\pm$ 0.11 & 56.19 $\pm$ 0.76 \\
        \hspace{3mm} Top-3 &    87.39\% (3.00) & 35.71\% $\pm$ 1.37 & 88.49\% $\pm$ 0.34 & 32.33 $\pm$ 0.17 & 9.84 $\pm$ 0.07 & 21.54 $\pm$ 0.13 & 59.19 $\pm$ 0.96 \\
        \hspace{3mm} Top-4 &    83.28\% (4.00) & 39.62\% $\pm$ 0.17 & 90.15\% $\pm$ 0.34 & 32.31 $\pm$ 0.21 & 9.69 $\pm$ 0.15 & 21.65 $\pm$ 0.15 & 59.86 $\pm$ 0.11 \\
        \hspace{3mm} Top-5 &    79.33\% (5.00) & 44.01\% $\pm$ 0.84 & 91.22\% $\pm$ 0.34 & 32.33 $\pm$ 0.24 & \textit{\textbf{9.84}} $\pm$ 0.21 & 21.67 $\pm$ 0.19 & 61.53 $\pm$ 0.35 \\
        \hspace{3mm} Top-6 &    75.35\% (6.00) & 44.60\% $\pm$ 0.89 & 92.65\% $\pm$ 0.17 & 32.55 $\pm$ 0.08 & 9.75 $\pm$ 0.21 & 21.39 $\pm$ 0.13 & 61.29 $\pm$ 0.46 \\
        \hline
        \multicolumn{7}{l}{Truncated Raw Content} \\
        \hline
        \hspace{3mm} First 6k words &    31.51\% (0.00) & 13.17\% $\pm$ 1.05 & 47.81\% $\pm$ 5.90 & 24.15 $\pm$ 1.42 & 4.89 $\pm$ 0.57 & 16.27 $\pm$ 0.96 & 61.43 $\pm$ 3.53 \\
        \hspace{3mm} Last 6k words  &    33.80\% (0.00) & 13.76\% $\pm$ 0.84 & 43.42\% $\pm$ 0.00 & 22.90 $\pm$ 0.10 & 4.35 $\pm$ 0.04 & 15.69 $\pm$ 0.03 & 52.47 $\pm$ 0.39 \\
        \hline
        \textbf{GistMem} &    82.81\% (0.00) & 44.96\% $\pm$ 0.44 & 91.93\% $\pm$ 0.73 & 31.20 $\pm$ 0.17 & 9.02 $\pm$ 0.09 & 20.60 $\pm$ 0.14 & 65.84 $\pm$ 0.87 \\
        \hline
        \multicolumn{7}{l}{{\textbf{ReadAgent-P}}} \\
        \hline
        \hspace{3mm} Look up 1 pg    &    79.37\% (0.98) & 44.84\% $\pm$ 0.00 & 92.29\% $\pm$ 0.34 & 31.46 $\pm$ 0.12 & 9.09 $\pm$ 0.11 & 20.63 $\pm$ 0.05 & 66.74 $\pm$ 0.74 \\
        \hspace{3mm} Look up 1-2 pgs &    77.00\% (1.72) & 43.42\% $\pm$ 1.01 & 92.88\% $\pm$ 1.05 & 31.77 $\pm$ 0.16 & 9.11 $\pm$ 0.12 & 20.70 $\pm$ 0.08 & 65.55 $\pm$ 0.28 \\
        \hspace{3mm} Look up 1-3 pgs &    74.85\% (2.46) & 44.37\% $\pm$ 1.21 & 91.22\% $\pm$ 0.44 & 31.89 $\pm$ 0.06 & 8.98 $\pm$ 0.13 & 20.70 $\pm$ 0.09 & 66.06 $\pm$ 1.63 \\
        \hspace{3mm} Look up 1-4 pgs &    73.26\% (3.02) & 44.13\% $\pm$ 0.50 & 90.51\% $\pm$ 0.44 & 31.87 $\pm$ 0.07 & 9.12 $\pm$ 0.06 & 20.77 $\pm$ 0.01 & 66.44 $\pm$ 0.74 \\
        \hspace{3mm} Look up 1-5 pgs &    72.01\% (3.44) & 43.42\% $\pm$ 1.45 & 91.22\% $\pm$ 0.60 & 31.80 $\pm$ 0.16 & 9.03 $\pm$ 0.07 & 20.64 $\pm$ 0.03 & 66.48 $\pm$ 0.39 \\
        \hspace{3mm} Look up 1-6 pgs &    70.65\% (3.89) & 42.70\% $\pm$ 1.54 & 90.51\% $\pm$ 0.73 & 31.74 $\pm$ 0.09 & 8.90 $\pm$ 0.09 & 20.66 $\pm$ 0.16 & 66.24 $\pm$ 1.14 \\
        \hline
        \textbf{ReadAgent-S} 1-6 pgs &    70.75\% (3.42) & \textbf{49.58\%} $\pm$ 0.44 & \textbf{93.83\%} $\pm$ 0.34 & \textbf{32.88} $\pm$ 0.15 & \textit{\textbf{9.98}} $\pm$ 0.06 & 21.50 $\pm$ 0.04 & 67.86 $\pm$ 0.11 \\
        \hline
    \end{tabular}
    \caption{
        \textbf{QMSum test} results (PaLM 2-L) means and standard deviations across 3 runs.
        35 articles and 281 questions.
        \textbf{Bold methods} are this work.
        \textbf{Bold} values are the best; \textit{\textbf{bold italics}} are ties for best.
        \textbf{CR} is the compression rate.
        \textbf{\# LU} is the number of lookups.
        \textbf{Resp. Length} is the length in words of the model's final response.
        We omit standard deviations for CR and \# LU for presentation purposes; they were all inconsequential.
    }
    \label{tab:qmsum_test}
\end{table*}